\documentclass{article} 
\usepackage{iclr2025_conference,times}
\usepackage{graphicx}
\usepackage{algorithm}
\usepackage{algpseudocode}

\usepackage{amsmath,amsfonts,bm}

\def\eqref#1{equation~\ref{#1}}

\def\1{\bm{1}}

\DeclareMathAlphabet{\mathsfit}{\encodingdefault}{\sfdefault}{m}{sl}
\SetMathAlphabet{\mathsfit}{bold}{\encodingdefault}{\sfdefault}{bx}{n}

\usepackage{hyperref}
\usepackage{url}
\usepackage{amsthm}
\usepackage{graphicx}        
\usepackage{pgfplots}        
\pgfplotsset{compat=1.18}    
\usepackage{subcaption}      

\makeatletter
\renewcommand{\ALG@beginalgorithmic}{\scriptsize} 
\makeatother

\iclrfinalcopy

\title{Recursive Self-Similarity in Deep Weight Spaces of Neural Architectures: A Fractal and Coarse Geometry Perspective}

\author{Ambarish Moharil, Indika Kumara, Damian Andrew Tamburri\\
Department of Data Governance, \\
The Jheronimus Academy of Data Science, \\
Eindhoven University of Technology \\
Eindhoven, The Netherlands \\
\texttt{\{a.moharil, i.p.k.weerasingha.dewage, d.a.tamburri\}}@tue.nl \\
\AND
Majid Mohammadi \\
The Vrije Universiteit Amsterdam \\
Amsterdam, The Netherlands \\
\texttt{majid.mohammadi690@gmail.com} 
\AND
Willem-Jan van den Heuvel \\
Department of Data Governance, \\
The Jheronimus Academy of Data Science, \\
Tilburg University \\
Tilburg, The Netherlands \\
\texttt{W.J.A.M.vdnHeuvel@tilburguniversity.edu} \\
}

\begin{document}

\maketitle

\begin{abstract}
This paper conceptualizes the Deep Weight Spaces (DWS) of neural architectures as hierarchical, \textit{fractal}-like, \textit{coarse} geometric structures observable at discrete integer scales through recursive dilation. We introduce a \textit{coarse group action} termed the fractal transformation, \( T_{r_k} \), acting under the symmetry group \( G = (\mathbb{Z}, +) \), to analyze neural parameter matrices or tensors, by segmenting the underlying discrete grid \( \Omega \) into \( N(r_k) \) fractals across varying observation scales \( r_k \). This perspective adopts a box count technique, commonly used to assess the hierarchical and scale-related geometry of physical structures, which has been extensively formalized under the topic of \textit{fractal geometry}. We assess the structural complexity of neural layers by estimating the \textit{Hausdorff-Besicovitch} dimension of their layers and evaluating a degree of self-similarity. The fractal transformation features key algebraic properties such as linearity, identity, and asymptotic invertibility, which is a signature of \textit{coarse structures}. We show that the coarse group action exhibits a set of symmetries such as \textit{Discrete Scale Invariance} (DSI) under recursive dilation, strong \textit{invariance} followed by weak \textit{equivariance} to permutations, alongside respecting the scaling equivariance of activation functions, defined by the intertwiner group relations. Our framework targets large-scale structural properties of DWS, deliberately overlooking minor inconsistencies to focus on significant geometric characteristics of neural networks. Experiments on CIFAR-10 using ResNet-18, VGG-16, and a custom CNN validate our approach, demonstrating effective fractal segmentation and structural analysis.
\end{abstract}

\section{Introduction}
\vspace{-2mm}
Until the 19th century, the term \emph{geometry} was \textit{synonymous} with Euclidean concepts. The pivotal works of Lobachevsky, Gauss, and Riemann later expanded the field by focusing on the study of non-Euclidean geometries. Subsequently, Felix Klein's Erlangen program redefined geometry as the study of \textit{invariances} under transformations \citep{geodesicbronstein}, which has been foundational to modern physical theories for studying conservation laws, emphasizing the importance of natural symmetries \citep{penrose2005reality}. This evolution underlies a recent trend in deep learning towards leveraging symmetries in understanding neural architectures \citep{geodesicbronstein}. Advances in deep learning have focused on treating the parameters of neural networks \textit{as a form of data}, encapsulating their inherent geometric properties and aligning with the Erlangen perspective of preserving \textit{symmetries} \citep{haggai}. This view facilitates the understanding and optimization of neural architectures given their geometric configurations \citep{battaglia2018relationalinductivebiasesdeep, eilertsen2020classifying, kalogeropoulos2024scaleequivariantgraphmetanetworks, haggai, navon2023equivariant, erkocc2023hyperdiffusion, metz2022velo, zhou2024permutation}. Moreover, recent research has demonstrated the efficacy of Graph Meta Networks (GMNs) that respect symmetries such as permutation invariance \citep{haggai, maron2018invariant} and scale equivariance \citep{kalogeropoulos2024scaleequivariantgraphmetanetworks, godfrey2022symmetries}. This underscores the importance of \textit{symmetries} in maintaining functional and geometric integrity when generating \textit{equivalent} representations of neural architectures, to improve their performance or interpretability \cite{maron2018invariant, haggai}.  Additionally, \cite{schürholt2024hyperrepresentationslearningpopulationsneural} et al. have emphasized the structured nature of the hyper-representation learning space (of neural architectures) while demonstrating that their weights encode latent properties such as the performance and training dynamics of the network, enabling model comparison across architectures, tasks, and training conditions.
 
In 1975, Benoit Mandelbrot introduced \emph{fractal geometry}, focusing on the complexity of natural forms by quantifying statistical \textit{self-similarity} and roughness through the \textit{Hausdorff-Besicovitch dimension} and emphasizing the significant impact of the \textit{observation scale} on measurements, which suggests that geometric measurements \textit{vary with observation} scale and are not \textit{truly} \textit{objective} \citep{Mandelbrot1978FractalsFC, mandelbrot1967how}. Despite advances in deep learning, a significant gap prevails in addressing the \textit{relativistic} perspective of \textit{geometric scale} when analyzing neural architectures’ Deep Weight Spaces (DWS) under recursive dilation. 

Recent works, such as \cite{WEN202187}, investigate self-similarity in single-layer static networks using various dimension metrics (e.g., fractal, information, and multi-fractal dimensions), highlighting advances in weighted and unweighted systems while emphasizing the unresolved fractality of multi-layer networks. Although research on multi-layered artificial networks is limited, \cite{Smith2021} examine the fractal geometry of biological neurons, demonstrating how the fractal dimension (\(D\)) integrates branch features (e.g., weave, length, width, tortuosity) to optimize connectivity while minimizing material, energy, and signal transmission costs. Additionally, \cite{Xue2017} argue that the study of \textit{fractality} and scaling behavior of complex networks requires the consideration of the interaction intensities or \textit{weights} along with a \textit{metric space}. They also demonstrate that the weights and metric spaces govern the existence of fractal properties, observable in both synthetic networks (e.g., Sierpinski triangles) and real-world systems (e.g., the human brain). These insights suggest that the application of fractal geometry and multifractal analysis to neural networks can enhance our understanding of their structural and functional complexities.

Motivated by the gaps in understanding fractal properties such as Discrete Scale Invariance (DSI)—a weaker form of Scale Invariance \citep{Sornette_1998} under recursive dilation, this study delves into the geometrical properties of neural architectures studied under coarse group operations over their DWS. We aim to view neural architectures as hierarchical fractal geometric structures that exhibit self-similarity across various scales. We focus on the structural and topological properties of DWS, which encode neural \textit{connectomics} in weight matrices or tensors. Thus, we propose a theoretical framework centered around a \textit{coarse action} \(T_{r_{k}}\), acting under the symmetry group \((\mathbb{Z},+)\) over some parameter grid $\Omega$. This transformation applies recursive isotropic discrete integer scaling over neural parameter signals, to reveal that certain integer dilation factors (\(\lambda\)) cause the Hausdorff-Besicovitch dimension (\(D\)) to exceed the topological dimension (\(D_T\)), indicating the fractal nature of the \textit{parameter signal geometry}. \(T_{r_{k}}\) is characterized by linearity, identity, and asymptotic invertibility, a \textit{signature of coarse structures} \cite{Roe2003CoarseGeometry, bunke2023coarsegeometry}, and demonstrates strong invariance under permutations, weak permutation equivariance, and strong discrete scale invariance, alongside traditional scale equivariance as defined by intertwiner group relations \citep{kalogeropoulos2024scaleequivariantgraphmetanetworks, godfrey2022symmetries}. These properties highlight the pivotal role of fractal transformations in elucidating the geometric complexities of neural architectures.
\vspace{-2mm}
\section{Preliminaries}
\vspace{-2mm}
\paragraph*{Fractal Geometry.}
The \textit{Euclidean dimension} \(E\) specifies the necessary coordinates in \(\mathbb{R}^n\), such as \(E = 1\) for lines, \(E = 2\) for planes, and \(E = 3\) for volumes. The \textit{topological dimension} \(D_T\), expanded upon by \cite{MengerUrysohn1922}, is the minimal dimension defining neighborhood boundaries. Further detailed by \cite{Hurewicz1941}, the Lebesgue covering dimension requires any open cover to be refined so no point overlaps in more than \(n+1\) sets. The Hausdorff–Besicovitch dimension, introduced by \cite{Hausdorff1919} and developed further by \cite{Besicovitch1938}, measures fractal space occupation and extends beyond integer dimensions, critical for complex structures like the Cantor set and the Koch curve  \citep{Cantor1883,Koch1904}.

Let \((\Omega, \rho)\) be a metric space, where \(\rho\) defines distances, and let \(S \subset \Omega\) be the subset of interest. The Hausdorff measure of \(S\) involves a \(\delta\)-cover, which is a collection of subsets \(\{U_i\}\) such that \(S \subset \bigcup_i U_i\) and \(\mathrm{diam}(U_i) \leq \delta\), where \(\mathrm{diam}(U_i) = \sup \{\rho(x, y) : x, y \in U_i\}\). For any real number \(d \geq 0\), the \(d\)-dimensional Hausdorff measure of \(S\) is defined as: $
\mathcal{H}^d(S) = \lim_{\delta \to 0} \inf \left\{ \sum_{i} \mathrm{diam}(U_i)^d : \{U_i\} \text{ is a } \delta\text{-cover of } S \right\}.
$ This measure considers coverings of \(S\) by balls (or subsets) of diameter \(\rho\), assigning the measure \(h(\rho) = \gamma(d) \rho^d\), where \(\gamma(d)\) is a constant dependent on \(d\), such as \(\gamma(d) = \frac{\pi^{d/2}}{\Gamma(1 + d/2)}\) for standard Euclidean spaces. The Hausdorff dimension of \(S\), denoted by \(D\), is the critical value of \(d\) where \(\mathcal{H}^d(S)\) transitions from infinite (\(d < D\)) to zero (\(d > D\)):
$D = \inf \{d \geq 0 : \mathcal{H}^d(S) = 0\}.$ For practical cases, the Hausdorff–Besicovitch dimension is often approximated using a logarithmic approach \textit{when \(S\) exhibits self-similarity}. If \(N(\rho)\) is the number of subsets of diameter \(\rho\) required to cover \(S\), the dimension can be computed as:
$D = \lim_{\rho \to 0} \frac{\log N(\rho)}{\log(1/\rho)}.$ 

Now, \emph{A fractal is defined as a set for which the Hausdroff-Besicovitch dimension strictly exceeds the topological dimension}\citep{Mandelbrot1978FractalsFC}.
For self-similar sets, such as the Cantor set \citep{Cantor1883} or the Koch curve \citep{Koch1904}, the fractal dimension is found to be \(D = \log(2)/\log(3) \approx 0.63\), and \(D = \log(4)/\log(3) \approx 1.26\) respectively \citep{Mandelbrot1978FractalsFC}.  

Additionally, fractal geometry intrinsically relates to \textit{scale invariance}, the property of a structure retaining self-similarity or statistical similarity under scale transformations. Formally, an observable \(\mathcal{O}(x)\) depending on a control parameter \(x\) is scale-invariant under \(x \rightarrow \lambda x\) if there exists a scaling factor \(\mu(\lambda)\) such that \(\mathcal{O}(x) = \mu \mathcal{O}(\lambda x)\). The solution to this equation takes the form of a power law, \(\mathcal{O}(x) = Cx^{\alpha}\), where \(\alpha = -\frac{\log(\mu)}{\log(\lambda)}\) \citep{Sornette_1998}. A key characteristic of scale invariance is that the ratio \(\frac{\mathcal{O}(\lambda x)}{\mathcal{O}(x)} = \lambda^{\alpha}\) remains independent of \(x\). \textit{Discrete Scale Invariance} (DSI) extends this concept, enforcing scale invariance only at specific integer-valued scales \(\lambda_n = \lambda_{n}.\lambda\), forming a countable set \citep{Sornette_1998,d7f62b04-88bf-3a3c-aad5-6ae5633ddd90}. Systems exhibiting DSI often display log-periodic oscillations in their scaling laws, which reveal discrete length or temporal scales governing their dynamics. DSI refines traditional scale invariance by identifying discrete organizational principles within systems, offering a more granular understanding of their structure and behavior \citep{Sornette_1998}.
\vspace{-2mm}
\paragraph*{Coarse Geometry.} \textit{Coarse geometry} is a mathematical framework developed to study the large-scale properties of spaces, focusing on relationships between points that persist under transformations, while disregarding small-scale variations. Introduced by J. Roe \citep{Roe2003CoarseGeometry}, coarse geometry formalizes this perspective by defining \textit{coarse spaces} \((X, C)\), where \(C\) is a collection of subsets of \(X \times X\) (coarse entourages) satisfying axioms of diagonal inclusion, closure under unions and subsets, symmetry, and compositionality. Coarse spaces generalize metric spaces by allowing infinite distances and encoding large-scale structures through coarse equivalences, which preserve these entourages. For a metric space \((X, d)\), coarse structures are induced by metric entourages \(U_r = \{(x, y) \in X \times X \mid d(x, y) < r\}\), and equivalences are defined by maps preserving such structures \citep{bunke2023coarsegeometry}. Coarse geometry finds applications in areas such as K-theory, topology, and mathematical physics, offering tools to study spaces that exhibit infinite-scale behaviors. Key concepts like \textit{coarse components}, \textit{Higson coronas}, and \textit{equivariant coarse homology theories} highlight its ability to encode and analyze large-scale invariants, often using interactions between metric, bornological, and coarse structures \citep{HigsonCoronasEquivariantHomology}.
\section{Related Works}
Recent advances in deep learning have focused on developing symmetry-aware metanetwork architectures that treat neural network parameters as training data \citep{haggai}. In this light, \cite{navon2023equivariant} propose Deep Weight-Space Networks (DWSNets), which leverage intrinsic permutation symmetries in Multilayer Perceptrons (MLPs), recognizing that permuting rows and columns of consecutive weight matrices preserves functional outputs. DWSNets utilize equivariant transformations to maintain these symmetries, demonstrating expressive capacity in approximating functions over weight spaces and excelling in tasks like domain adaptation, classification of implicit neural representations (INRs) followed by weight-space analysis. Moreover, it has been shown that ignoring the above symmetries underlying DWS leads to the degradation of metanetwork performance \citep{peebles2022learninglearngenerativemodels, navon2023equivariant}

Expanding on this, \citep{maron2018invariant, haggai} have worked on universal graph meta-network architectures that remain invariant or equivariant to permutations of neural nodes, enabling learning of neural representations in deep weight spaces. Moreover, \cite{zhou2024permutation} developed Neural Functional Networks (NFNs), which embed permutation symmetries of feedforward networks directly into their architecture. They show that NF-Layers in NFNs maintain equivariance or invariance under hidden and full neuron permutations, aggregate weight-space features into permutation-invariant scalars, and approximate symmetric functions over neural weight spaces. This approach supports applications like classifier generalization prediction, sparse training mask generation, and weight editing for style transformation while reducing reliance on augmentation, improving computational efficiency and generalization.

Building on the foundations, \cite{godfrey2022symmetries} formalize intertwiner groups \( G_\sigma^n \) as symmetries of nonlinear activation functions, defined as the set of invertible linear transformations on the hidden activation space that commute with the activation \(\sigma\). This links weight space symmetries and invariant realization maps in function space, unifying the understanding of permutation symmetry and scaling invariance in neural networks, particularly in architectures with batch normalization. Building on this, \cite{kalogeropoulos2024scaleequivariantgraphmetanetworks} extend the framework to scaling symmetries in activations like ReLU and tanh, characterized by transformations \( \sigma(ax) = b\sigma(x) \), influencing both weight space and data representations. To address these symmetries, they propose \textit{ScaleGMNs}, which integrate scaling equivariance and permutation symmetry using rescale-equivariant message-passing layers, ensuring that the vertex and edge representations in metanetworks respect these transformations while embedding inductive biases directly into neural architectures.
\vspace{-2mm}
\section{Fractal Geometry over Deep Weight Spaces}\label{secs:fracdws}
\vspace{-2mm}
Let \(W \in \mathbb{R}^{m \times n}\) be a matrix where \(m\) and \(n\) are positive integers. This matrix represents a real-valued signal defined on a finite integer grid. The grid is given by: $\Omega = \{(i, j) \mid 1 \leq i \leq m \text{ and } 1 \leq j \leq n\} \subset \mathbb{Z}^2.
$ Each point \((i, j) \in \Omega\) corresponds to an entry \(W_{i,j} \in \mathbb{R}\). This allows \(W\) to be interpreted as a function  $ W: \Omega \to \mathbb{R} \quad \text{where} \quad (i, j) \mapsto W_{i,j}.$  

The grid \(\Omega\) and the values associated with it reside in a three-dimensional Euclidean space, where each point in matrix \(W\) is uniquely determined by the coordinates \((i, j, W_{i,j})\) in \(\mathbb{R}^3\). The indices \((i, j)\) lie within \(\mathbb{R}^2\), and \(W_{i,j}\) represents the signal value at these indices, necessitating a Euclidean dimension \(E = 3\) for embedding. Topologically, \(\Omega\) is viewed as a two-dimensional integer grid analogous to a two-dimensional surface, assigning it a topological dimension \(D_T = 2\), indicating the minimal parameters required to describe local neighborhoods \citep{MengerUrysohn1922}. This dimensionality is crucial for understanding the matrix \(W\) defined on this grid. Additionally, for examining \(\Omega\)'s large-scale attributes, we employ a coarse geometry framework paired with fractal geometry principles. In this setting, \(\Omega\) adopts a metric coarse structure defined by the \(\ell^1\)-distance, which efficiently handles the small, irregular boundary subsets as bounded but negligible in influencing the grid’s broader behavior \citep{Roe2003CoarseGeometry, bunke2023coarsegeometry}.


In the \textit{coarse} geometric framework we endow $\Omega$ with an $\ell^{1}$ metric $d(\cdot,\cdot)$:
\[
d\bigl((i_1,j_1),(i_2,j_2)\bigr) \;=\; \tfrac{1}{2}\,\Bigl(
\lvert i_1-i_2\rvert \;+\;
\lvert j_1-j_2\rvert
\Bigr).
\]
Here, an $\ell^1$-distance effectively corresponds to a \textit{square tile} of side $r_{k}$ \cite{Mandelbrot1978FractalsFC, Roe2003CoarseGeometry}. \footnote{ Dividing by 2 is convenient: if $d(x,y)\le r$, then $\lvert i_1-i_2\rvert+\lvert j_1-j_2\rvert\le 2r$. This unifies the notion of ``block size'' and ``distance threshold'' in $\Omega$.} The distance function serves two purposes; by setting $d(x,y) \leq r_{k}$ we capture the large scale relationships i.e the \textit{entourages}. Additionally, by setting $d(x,y) < r_{k}$, we capture the local, small-scale boundedness.
To study the large-scale relationships of the signal $W$ and inturn the grid $\Omega$ we define \textit{coarse} \textit{entourages} as subsets of $\Omega \times \Omega$ containing pairs $((i_{1},j_{1}),(i_{2},j_{2}))$ seperated by atmost $r_{k}$ as:
\[
E_{r_k} \;=\; \bigl\{
\bigl((i_1,j_1),(i_2,j_2)\bigr)\in \Omega\times\Omega
\;\bigm|\;
d((i_1,j_1),(i_2,j_2))\;\le\; r_k
\bigr\}.
\]
The \emph{coarse structure} $C$ is the \textit{family} of all finite unions of $E_{r_k}$, closed under subsets, inversions $(x,y)\mapsto(y,x)$, and composition; $E_1\circ E_2 \;=\; \{(x,z)\mid \exists\,y:\,(x,y)\in E_1,\,(y,z)\in E_2\}.$
subject to, the diagonal $\Delta=\{(x,x)\mid x\in\Omega\}$ trivially sits in every $E_{r_k}$ once $r_k>0$. If $E_{r_{k}}$ and $E_{r_{l}}$ are entourages across scales $r_{k}$ and $r_{l}$ respectively, then so is their union, and so on (\textbf{Proposition 1}). And being a finite grid, once $r_k$ exceeds half the diameter, $E_{r_k}=\Omega\times\Omega$ \citep{bunke2023coarsegeometry}. This structure captures the global connectivity of $\Omega$ while ignoring the \textit{small}-\textit{scale} irregularities. Each entourage $E_{rk}$ encodes pairs $(x,y)$ exasperated by at most $r_{k}$ in the $\ell^{1}-$distance. For sufficiently large $r_{k}$ the entourages $E_{r_{k}}$ can encompass essentially all of $\Omega \times \Omega$. 

Within our framework, we define the covering by a set of \textit{isometric} squares of size $r_{k}$ and thus we define the set of local open \textit{tilings} as:
$$B_{\text{local}}(x,r_{k}) = \{y \in \Omega \mid d(x,y) < r_{k}\}$$

Hence, a subset $A\subset \Omega$ is \emph{bounded} if it fits in some open \textit{tile} of scale $r_{k}>0$ (\textbf{Proposition 4} \ref{proofs:prop4}):
$
A\;\subset\;
B_{\mathrm{local}}(x,r_{k})
$. The collection $B$ of all bounded subsets yields the \textit{Bornological} structure $(\Omega,B)$ \citep{Roe2003CoarseGeometry, bunke2023coarsegeometry}. By requiring $d(x,y)<r_{k}$, we focus on truly small-scale neighborhoods. Hence, a set is bounded if it lies in some open ball or \textit{tile} (under $\ell^{1}$-distance of strictly less than radius $r_{k}$).\footnote{In a finite grid we have that every finite subset is trivially bounded yet this definition helps us systematically track local aspects of fractal transformations.}

To model scale-dependent transformations, we define a group \(G = (\mathbb{Z}, +)\) where addition is performed as \( (k, \ell) \mapsto k + \ell \) for all \( k, \ell \in \mathbb{Z} \). Each \( k \) denotes a scale \( r_k \) formulated as \( r_k = \left\lfloor \frac{r_0}{\lambda^k} \right\rfloor \) with \( r_0 = \min(m, n) \) and \( \lambda > 1 \), representing increasingly finer partitions of \(\Omega\) as \( k \to \infty \). For any matrix \( W \in \mathbb{R}^{m \times n} \) defined on the integer grid \(\Omega \subset \mathbb{Z}^2\), the coarse group action \( T_{r_k} \) under \( G \) employs dilation parameter \( \lambda \) to determine the observation scale \( r_k \) at each step \( k \), analyzing neural connections at various layers by decomposing the signal into $N(r_{k})$ fractals. The integer grid’s regular but restrictive spacing necessitates integer scale segmentations, potentially causing discontinuities in tiling when \( m \) or \( n \) is not divisible by \( r_k \), with \( W \) segmented into \( r_k \times r_k \) blocks through recursive application of \( T_{r_k} \). Specifically, signal \(W\) is partitioned into $r_{k} \times r_{k}$ blocks by recursive application of the \textit{corase group action} $T_{r_{k}}: \Omega \mapsto\Omega$ at respective scales $r_{k}$ as:
\[
T_{r_k}(W)
\;=\;
\bigcup_{(p,q)\,\in\, \mathcal{I}_{r_k}}
\,
P_{(p)}\,W\,Q_{(q)}^\top.
\]
For each top-left index \((p,q)\) and scale \(r_k\) we define \textit{selector} matrices \(P_{(p)}\) and \(Q_{(q)}\) that \textit{extract} row and column subsets. These blocks may partially exceed \(\Omega\)’s boundary but remain finite in a finite grid. Some edges might lead to smaller blocks if \(p+r_k>m\) or \(q+r_k>n\). 

\textbf{Proposition 1.}\label{proofs:prop1}
For any \(k, \ell \in G = (\mathbb{Z}, +)\) (\textbf{Theorem 0} \ref{proofs:thmZgroup}), the composition of transformations satisfies $T_{r_k} \circ T_{r_\ell} = T_{r_{k+\ell}}$. This property ensures that the family of coarse group\textit{ transformations  \(\{T_{r_k}\}_{k \in (\mathbb{Z,+})}\) remain closed under composition.} (Proof of \textbf{Proposition 1}  \ref{proofs:prop1}, Appendix \ref{proofs})  

This captures how one can “\textit{first do scale \(\ell\), then do scale \(k\),}” which is the same as “\textit{do scale \(k+\ell\).}” It is the \textit{\textbf{essence of scaling or iterative refinement in fractal or wavelet applications}} \cite{Sornette_1998}.

Moreover, we establish the linearity of the family of fractal transformations $\{T_{r_{k}}\}_{k \in (\mathbb{Z},+)}$ in \textbf{Proposition 2} \ref{proofs:prop2}, the identity property in \textbf{Proposition 3} \ref{proofs:prop3}, coarse geometric property of boundedness in \textbf{Proposition 4} \ref{proofs:prop4}, large-scale uniformity of  $\{T_{r_{k}}\}_{k \in (\mathbb{Z},+)}$ in \textbf{Proposition 5} \ref{proofs:prop5} and the \textit{coarse equivalence} in \textbf{Proposition 6} \ref{proofs:prop6}.

\textbf{Theorem 7.}\label{proofs:prop7} \textit{The family of transformations \(\{T_{r_k}\}_{k \in (\mathbb{Z,+})}\) ( $\Omega \mapsto \Omega$) forms a coarse group action under \((\mathbb{Z}, +)\) on $(i,j) \in \Omega$). Specifically: 1. The transformations are coarsely proper (Proposition \ref{proofs:prop4}). 2. The action is cobounded, with \(\Omega = \bigcup_{k \in \mathbb{Z}} T_{r_k}(U)\) for some bounded \(U \subseteq \Omega\) (Proposition \ref{proofs:prop4}). 3 .The transformations are large-scale uniform (Proposition 5 \ref{proofs:prop5}). The composition satisfies the additive property: $T_{r_k} \circ T_{r_\ell} = T_{r_{k+\ell}},$ which ensures consistency with the group structure. By Corollary 6.2 of \cite{N2008}, this implies that the action is coarse, and \((\mathbb{Z}, C_\mathbb{Z})\) is coarsely equivalent to \((\Omega, C_\Omega)\).} (Proof of \textbf{Theorem 7} \ref{proofs:prop7}).


\textbf{Theorem 8.}\textit{
The matrix signal \( W \in \mathbb{R}^{m \times n}\) exhibits \textit{\textbf{discrete scale-invariance (DSI)}} if there exists a constant \( D \) (the fractal dimension) such that: $N(r_k) \sim \lambda^{D} N(r_{k+1}) \quad \text{as } k \to \infty.$ If: $m \bmod r_k = 0 \quad \text{and} \quad n \bmod r_k = 0 \quad \text{for all } k \geq 0,$ then the submatrices perfectly tile \( W \), and: $N(r_k) = \lambda^{2k},$
implying \( D= 2 \), which matches the topological dimension $D_{T} = 2$ of the grid \( \Omega \). If there exists any \( k \geq 0 \) such that: $m \bmod r_k \neq 0 \quad \text{or} \quad n \bmod r_k \neq 0,$
then: $N(r_k) > \lambda^{2k},$ indicating \( D > 2 \). The presence of incomplete subdivisions at certain scales increases the covering count \( N(r_k) \) beyond the scaling factor \( \lambda^{2k} \), revealing fractal behavior in the geometry of \( W \).} 

Fractal analysis of matrix \(W\) on grid \(\Omega \subset \mathbb{Z}^2\) evaluates divisibility of dimensions \(m\) and \(n\) by scale \(r_k\). Perfect divisibility across all scales results in exact tiling, with \(N(r_k) = \lambda^{2k}\) indicating a Euclidean structure as fractal dimension \(D\) matches topological dimension \(D_T = 2\). If divisibility fails, residuals inflate \(N(r_k)\) to \(C \cdot (\lambda^{2+\epsilon})^k\), elevating \(D\) above \(D_T\), illustrating Mandelbrot's scale-invariance and complex pattern interplay (see Appendix \ref{proofs:prop8} for proof).

To contextualize discussions on DWS and group actions, we detail $T_{r_{k}}$'s coarse geometric properties in Appendix \ref{proofs}, with \textbf{Theorem 8} \ref{proofs:prop8} confirming its structural invariance under recursive dilation. We also examine permutation invariance and equivariance, essential in neural architecture studies \cite{HECHTNIELSEN1990129, haggai}, considering how a permutation operator $P_\pi \in GL(n,\mathbb{R})$ interacts with $T_{r_{k}}$ and impacts neural meta-network architectures.

\textbf{Proposition 9.}\label{proofs:prop9} \textit{Given a matrix \( W \) in \( \mathbb{R}^{m \times n} \) and permutation matrices \( P_\pi^r \in \mathbb{R}^{m \times m} \) for rows and \( P_\pi^c \in \mathbb{R}^{n \times n} \) for columns, the coarse group action \( T_{r_k} \) maintains global structural invariance such that the overall segment count \( N(r_k) \) remains consistent across scales regardless of permutations: \( W \cdot P_\pi \cdot T_{r_k} \underset{C}{\sim} W \cdot T_{r_k} \cdot P_\pi \). However, at the level of individual submatrices, \( T_{r_k} \) exhibits weak equivariance where local compositions may differ based on the permutations, reflecting adjustments in grid-based segmentation: \( T_{r_k} \cdot W \overset{\approx}{\to} T_{r_k} \cdot (P_\pi^r \cdot W \cdot P_\pi^c) \). This highlights how the coarse geometry framework adapts to permutations, preserving overall structure while allowing flexibility at the local level.} (Proof \ref{proofs:prop9})

The above proposition underscores that the fractal transformation \( T_{r_k} \) does not inherently adjust to node permutations within neural networks. Permutations alter the matrix \( W \)'s state but not its content, necessitating a reinitialization of the segmentation to adapt to this new state. The fractal segmentation process, deterministic in nature, starts from the top-left index and segments \( W \) into \( r_k \times r_k \) submatrices based on the grid structure and scale \( r_k \), independent of the matrix's internal value arrangement. Post-permutation, the matrix requires new observations on this altered state, acknowledging that while global structural properties remain invariant, local content variations may require resetting the observation mechanism. This geometric characteristic of \( \{T_{r_{k}}\}_{k\in (\mathbb{Z},+)} \) ensures the continued relevance of fractal insights into the altered weight space of neural architectures.

Now, shifting our focus on the functional symmetries as studied in the works \cite{godfrey2022symmetries, kalogeropoulos2024scaleequivariantgraphmetanetworks}, we say: 
A neural network \( f : \mathbb{R}^{n_0} \to \mathbb{R}^{n_k} \) consists of \( k \) layers, each being an affine transformation followed by a nonlinearity \( \sigma \), except the final layer which omits \( \sigma \). For each layer \( i \):$ \ell_i(x) = 
\begin{cases} \sigma(W^i x + b^i), & \text{if } i < k, \\W^i x + b^i, & \text{if } i = k,\end{cases} $
where \( W^i \) and \( b^i \) are the weights and biases respectively. The network function \( f \) composes these layers: \( f = \ell_k \circ \cdots \circ \ell_1 \). Applying the fractal transformation \( T_{r_k} \) to an activation map \( h^i \), where \( h^i = \sigma(W^i h^{i-1} + b^i) \) and \( h^0 = x \), segments \( h^i \) into \( N(r_k) \) fractal activations via: $ T_{r_k}(h^i) = \bigcup_{(p, q) \in \mathcal{I}_{r_k}} \sigma\left(P_p W^i Q_q^\top x + P_p b^i\right) $ Here, \( P_p \) and \( Q_q \) are selector matrices determining the fractal segment \( h_{p, q}^{i, r_k} \) within the grid-based segmentation of \( h^i \).

\textbf{Proposition 10.}\label{proofs:prop10} \textit{Each localized fractal activation \( h_{p, q}^{i, r_k} \) corresponds to a specific region in the activation map of the \( i \)-th layer, determined by the selector matrices \( P_p \) and \( Q_q \). It captures the contribution of the submatrix \( P_p W^i Q_q^\top \) to the layer’s output within the \( r_k \times r_k \) neighborhood indexed by \( (p, q) \). This decomposition reflects the localized and self-similar structure of the activation map across scales \( r_k \).} (Proof \ref{proofs:prop10}.)

having proposed this, we also focus on understanding the symmetries inherently present within the activation functions of a Feed Forward Neural Network (FFNN), specifically where there exist pairs $(a, b)$ for which it holds that $\sigma(ax) = b\sigma(x)$, more formally defined as the intertwiner group \citep{godfrey2022symmetries}. In this context, we make the following proposition:

\textbf{Proposition 11.}\label{proofs:prop11} \textit{Let \( \sigma: \mathbb{R} \to \mathbb{R} \) be a continuous, non-linear activation function with an intertwiner group \( G_{\sigma_{n}} = \{A \in \mathbb{R}^{n_{i} \times n_{i}} : \text{invertible} \, | \, \exists B \in \mathbb{R}^{n_{i-1}\times n_{i-1}} : \text{invertible}, \, \sigma(Ax) = B\sigma(x) \} \). For any layer \( i \), let \( A \in GL_{n_{i}}(\mathbb{R} \) and \( B \in GL_{n_{i-1}}(\mathbb{R}) \). Then, under the fractal transformation \( T_{r_k} \), the activation map \( h^i \) and its fractals \( h_{p, q}^{i, r_k} \) satisfy the scaling relationship: $\sigma_{n_i}\left(A \widetilde{W}_{(p, q)}^{i, r_k} x + Ab^i\right) = B \sigma_{n_i}\left(\widetilde{W}_{(p, q)}^{i, r_k} x + b^i\right),$where \( \widetilde{W}_{(p, q)}^{i, r_k} \) is the lifted fractal matrix. This ensures that the scaling relationship holds globally and fractally, consistent with the intertwiner properties of \( A \) and \( B \).} (Proof \ref{proofs:prop11}.)

The fractal transformation, under \( (\mathbb{Z}, +) \), preserves intertwiner group relations, ensuring that local computations align with global representations. This guarantees predictable behavior under scaling and linear transformations, maintaining the integrity of relationships within the matrix. It provides a robust framework for analyzing multiscale patterns and structural invariance, facilitating an understanding of hierarchical relationships within activation maps \cite{godfrey2022symmetries}.

When processing multi-channel images like RGB, each channel \(C_1, C_2, C_3\) is convolved by a kernel \(K_i\) to capture specific color patterns, generating feature maps combined via summation and non-linearity \(\sigma\). Consider a 4D weight tensor \(W\) with shape \([N, C, k_1, k_2]\), where \(N\) is the neuron count, \(C\) the channel count, and \(k_1, k_2\) the kernel dimensions, with \(W_{i,j} \in \mathbb{R}^{k_1 \times k_2}\) as the kernel for neuron \(i\) and channel \(j\) on the left and then on the right \(W_{i,j}\) which contains \textit{sub-tensors} \(W_{i,j}^{p,q} \in \mathbb{R}^{k'_1 \times k'_2}\), represented as:
$$
W = \left[
\begin{array}{cccc}
W_{1,1} & W_{1,2} & \cdots & W_{1,C} \\
W_{2,1} & W_{2,2} & \cdots & W_{2,C} \\
\vdots & \vdots & \ddots & \vdots \\
W_{N,1} & W_{N,2} & \cdots & W_{N,C}
\end{array}
\right],
\quad
W_{i,j} = \left[
\begin{array}{cccc}
W_{i,j}^{1,1} & W_{i,j}^{1,2} & \cdots & W_{i,j}^{1,k_2} \\
W_{i,j}^{2,1} & W_{i,j}^{2,2} & \cdots & W_{i,j}^{2,k_2} \\
\vdots & \vdots & \ddots & \vdots \\
W_{i,j}^{k_1,1} & W_{i,j}^{k_1,2} & \cdots & W_{i,j}^{k_1,k_2}
\end{array}
\right],
$$
capturing multi-scale structural details, analogous to the Tensor-Train format \citep{oseledets2011tensor, gelss2018tensor}. Applying fractal segmentation via coarse group action \(T_{r_k}\) under \( (\mathbb{Z}, +) \), \(W\) is segmented into; $T_{r_k}(W) = \bigcup_{(p, q) \in \mathcal{I}_{r_k}} P_p W Q_q^\top,$ where \( P_p \in \mathbb{R}^{N \times N} \) and \( Q_q \in \mathbb{R}^{C \times C} \) are selector matrices for neuron and channel segments, generating \(N(r_k)\) fractal segments at scale \(r_k\), each reflecting localized patterns (Algorithm \ref{algo:fractal_segmentation_4d}).
\vspace{-2mm}
\section{Experiments}
\begin{figure}[t!]
    \centering
    \begin{subfigure}[b]{0.5\textwidth}
        \centering
        \includegraphics[width=\textwidth]{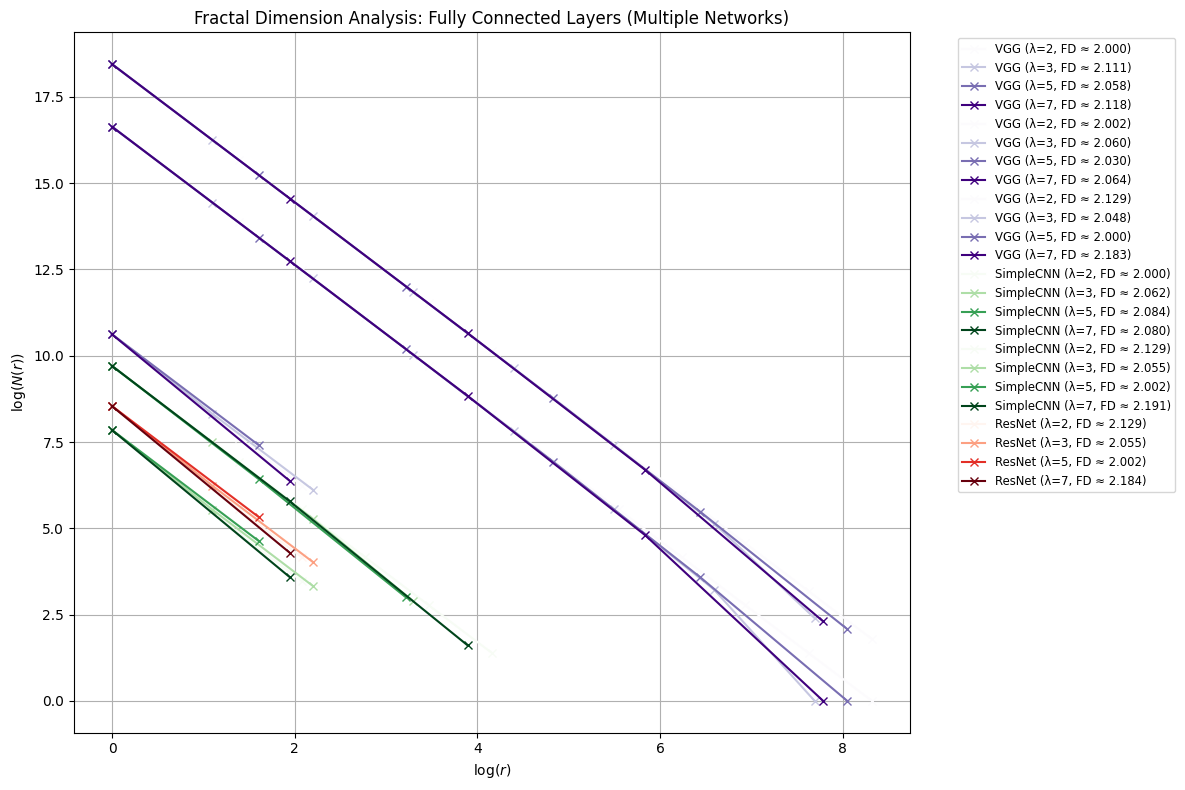} 
        \caption{}
    \end{subfigure}
    \hfill
    \begin{subfigure}[b]{0.44\textwidth}
        \centering
        \includegraphics[width=\textwidth]{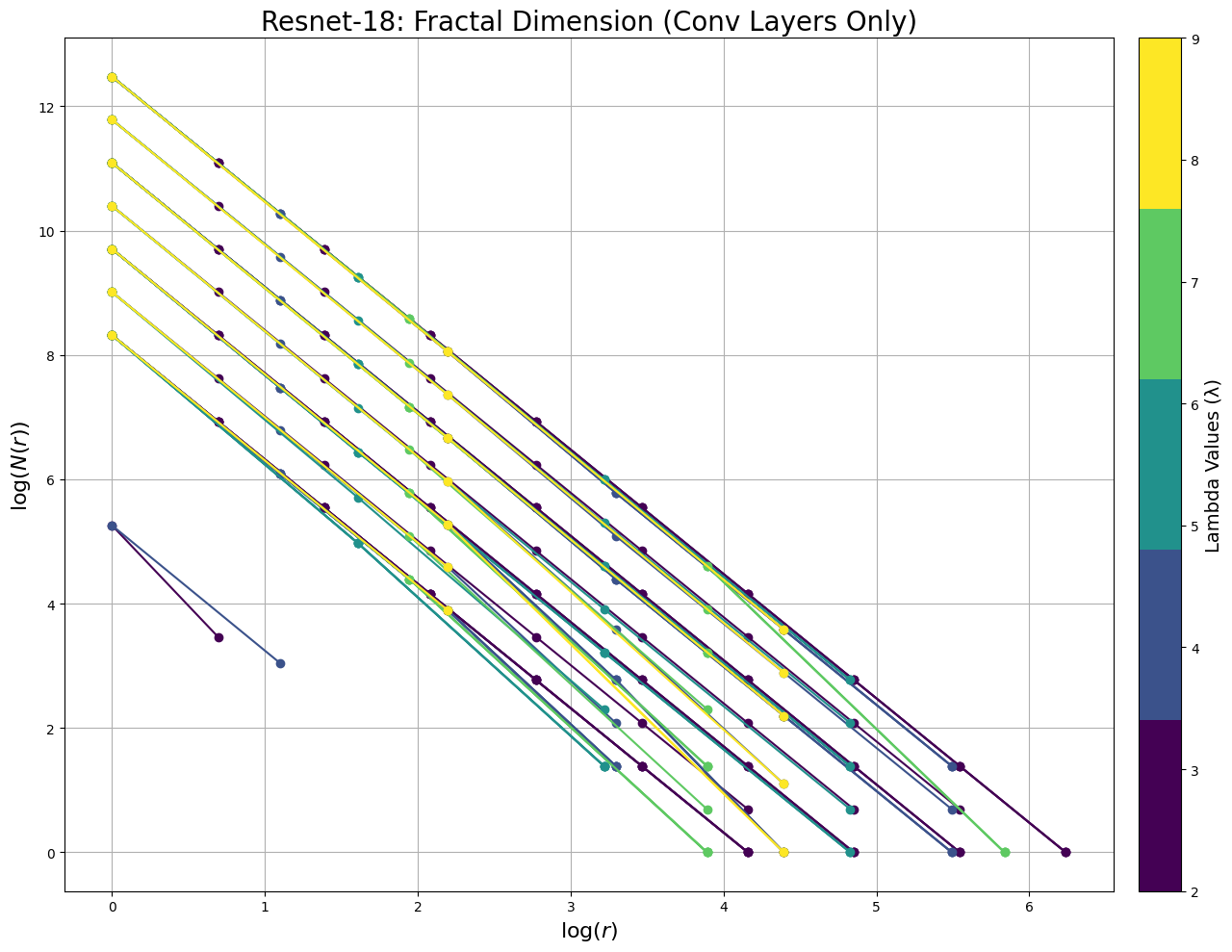} 
        \caption{}
    \end{subfigure}
    \vspace{-2mm}
\caption{
The top-left figure (a) shows the log-log plot for estimating the fractal dimension \(D\) of the fully connected layers in ResNet-18, VGG-16, and SimpleCNN, analyzed across \(\lambda = \{2, 3, 5, 7\}\). The top-right figure (b) depicts the \(D\) estimation for ResNet-18’s convolutional layers for \(\lambda = \{2, 3, 5, 7, 9\}\). Consistent \(D\) values across scales indicate self-similarity and structural invariance, with convergence towards \(D = 2\) reflecting hierarchical abstraction in deeper layers.}
\label{figexp}
\vspace{-5mm}
\end{figure}

\begin{figure}[h!]
    \centering
    \begin{subfigure}[b]{0.48\textwidth}
        \centering
        \includegraphics[width=\textwidth]{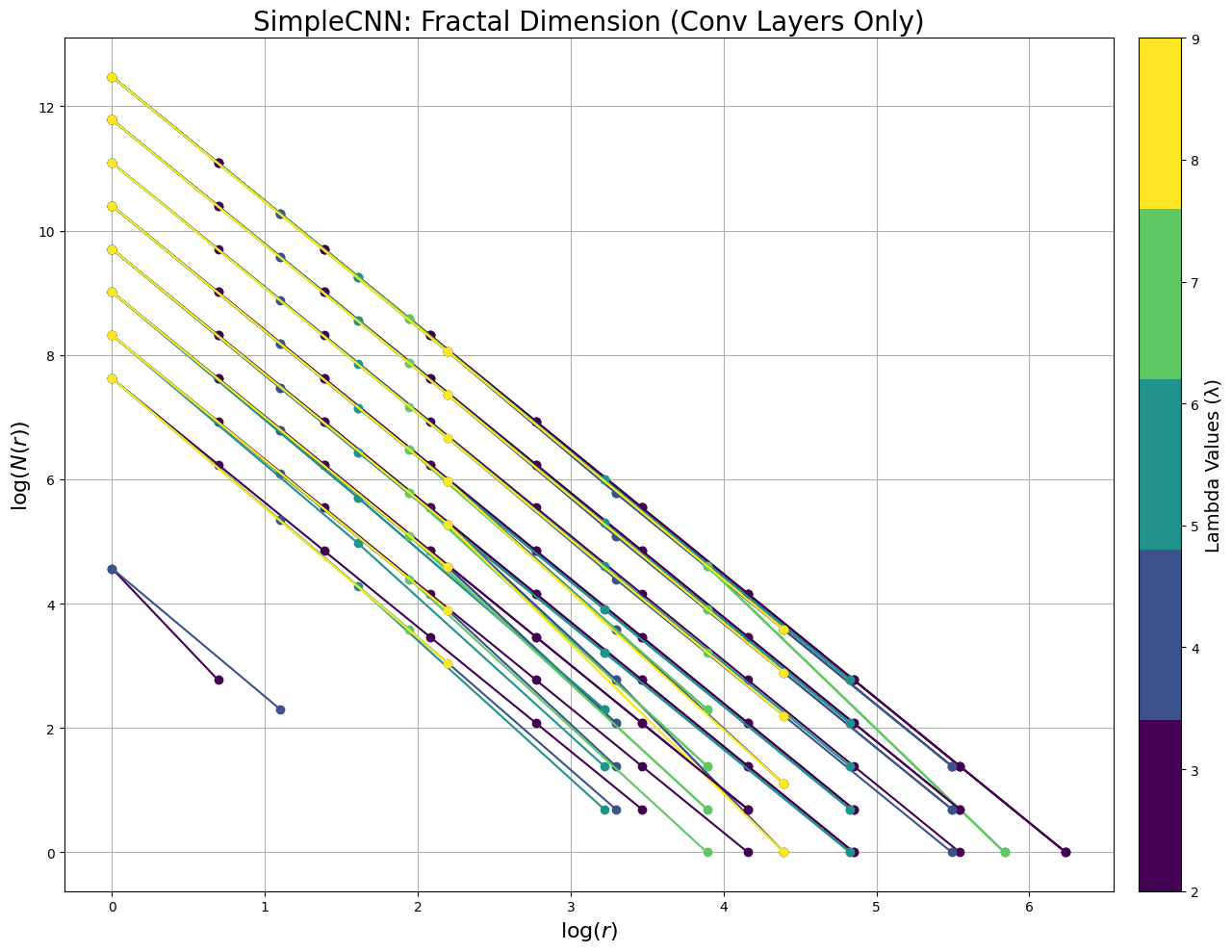} 
        \caption{}
    \end{subfigure}
    \hfill
    \begin{subfigure}[b]{0.48\textwidth}
        \centering
        \includegraphics[width=\textwidth]{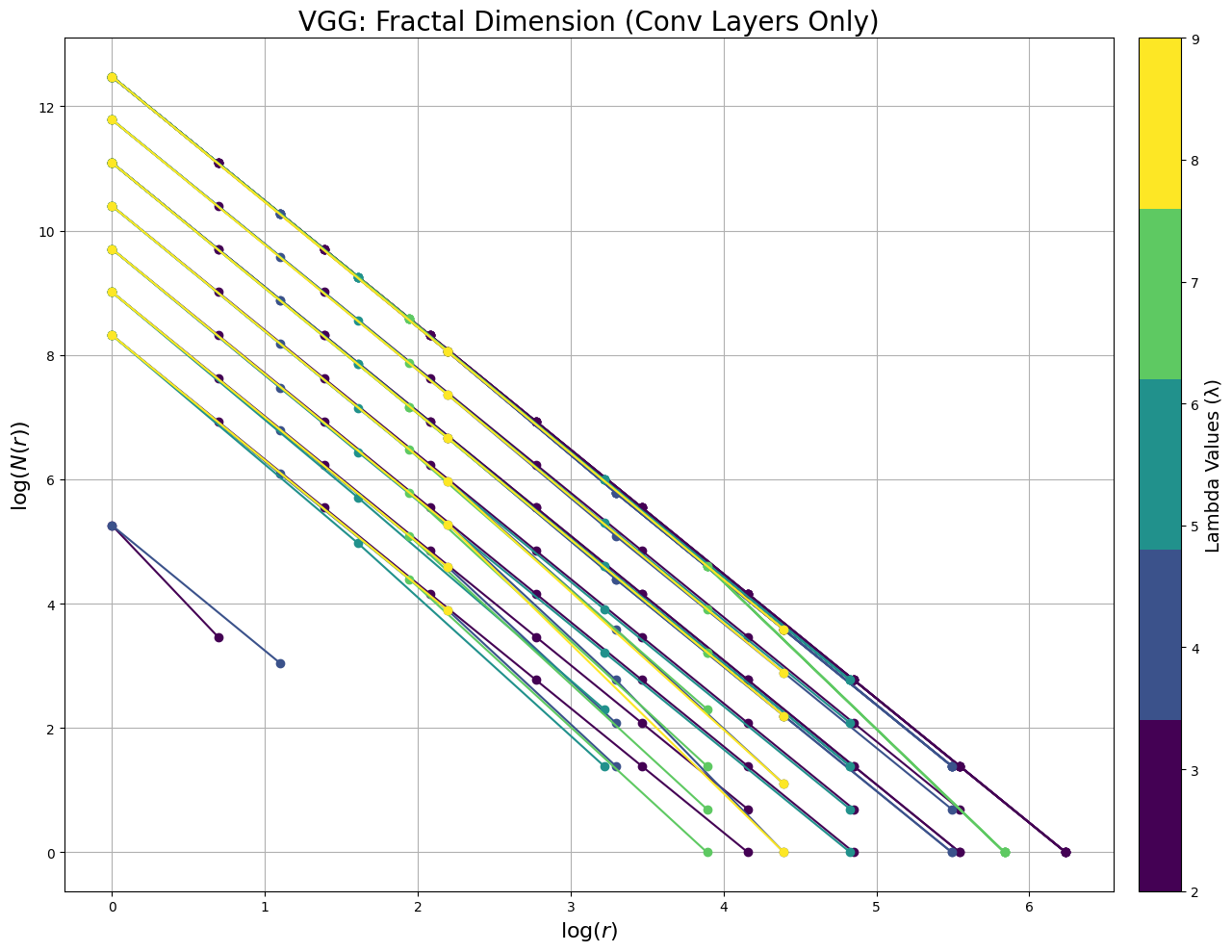} 
        \caption{}
    \end{subfigure}
\caption{
 plots (a) and (b) show the \(D\) estimations for SimpleCNN and VGG-16, respectively. Consistent \(D\) values across scales indicate self-similarity and structural invariance, with convergence towards \(D = 2\) reflecting hierarchical abstraction in deeper layers.}
\label{figexp2}
\vspace{-5mm}
\end{figure}

We analyzed ResNet-18, VGG-16, and a 15-layer SimpleCNN using the fractal transform \(T_{r_k}\) to estimate layer-wise fractal dimensions (\(D\)) across scales (\(\lambda = 2, 3, 5, 7, 9\)), revealing structural consistency and transitions from localized to global abstractions (Fig. \ref{figexp} \& Fig \ref{figexp2} \ref{add_res}). All the architectures exhibit distinct fractal behaviors reflecting their respective architectural designs and hierarchical feature representations, while sharing structural self-similarity under recursive dilation (Algorithms \ref{algo:fractal_segmentation}, \ref{algo:fractal_segmentation_4d}, \ref{algo:fd_analysis_conv}, \ref{algo:fd_analysis_multiple_fc}).

Initial convolutional layers in all architectures exhibit high fractal complexity (\(D_{\lambda=2} = 2.5850\)), indicating robust localized feature extraction, but \(D\) values decrease sharply at larger scales (\(D_{\lambda=3} = 2.0143\)), reflecting kernel size constraints. In ResNet-18, early layers (\texttt{layer1.0.conv1.weight}, \texttt{layer1.1.conv1.weight}) maintain consistent fractal dimensions (\(D_{\lambda=3} = 2.0928\), \(D_{\lambda=5} = 2.1534\)), a hallmark of its residual connections, which ensure periodic structural coherence while facilitating transitions toward broader abstractions. The middle residual blocks (\texttt{layer2.x}) in ResNet-18 exhibit increasing complexity across scales, with \(D_{\lambda=7} = 2.1372\) and peaks such as \(D_{\lambda=9} = 2.2083\) in \texttt{layer2.0.conv2.weight}, signifying the network's ability to capture broader spatial abstractions. This trend continues in the deeper layers (\texttt{layer3.x}, \texttt{layer4.x}), where fractal dimensions stabilize (\(D_{\lambda=3} \approx 2.0230\), \(D_{\lambda=5} = 2.0113\)) while occasional peaks (\(D_{\lambda=7} = 2.1278\) in \texttt{layer4.0.conv2.weight}) indicate sensitivity to larger-scale patterns. These fluctuations demonstrate how ResNet-18 dynamically balances localized and global feature abstractions across layers.

In contrast, VGG-16’s early layers (\texttt{features.2.weight}, \texttt{features.5.weight}) share similarities with ResNet-18, maintaining consistent fractal dimensions (\(D_{\lambda=3} = 2.0928\), \(D_{\lambda=5} = 2.1534\)) and reflecting uniformity at intermediate scales. However, middle convolutional layers (\texttt{features.5.weight}, \texttt{features.7.weight}) show a divergence, with fractal dimensions peaking at \(D_{\lambda=9} = 2.2083\), reflecting a greater emphasis on progressive spatial abstraction. Deeper layers (\texttt{features.17.weight} to \texttt{features.28.weight}) stabilize (\(D_{\lambda=3} = 2.0230\), \(D_{\lambda=5} = 2.0113\)) with periodic structural shifts, highlighting refinement of global patterns while maintaining recursive dimensional similarity. SimpleCNN follows a more predictable fractal profile due to its minimalistic architecture. Middle layers (\texttt{conv3.weight} to \texttt{conv8.weight}) stabilize around \(D_{\lambda=3} \approx 2.0928\), reflecting structural self-similarity akin to ResNet-18 and VGG-16. However, broader patterns emerge in \texttt{conv4.weight}, peaking at \(D_{\lambda=9} = 2.2083\), while deeper layers (\texttt{conv9.weight} to \texttt{conv15.weight}) converge around \(D \approx 2.02\), consolidating high-level abstractions. When \(D\) rises across \(\lambda\) in one layer but declines in the next, SimpleCNN alternates between broader pattern abstraction and refocusing on localized features, illustrating a simpler yet effective hierarchical progression.
Fully connected layers across the three architectures further reveal their contrasting approaches to feature abstraction. ResNet-18’s \texttt{fc.weight} transitions dynamically, starting at \(D_{\lambda=2} = 2.1288\), dropping to \(D_{\lambda=5} = 2.0024\), and rising at \(D_{\lambda=7} = 2.1843\), balancing feature consolidation and sensitivity to broader patterns. VGG-16’s final fully connected layer mirrors this trajectory, peaking sharply at \(D_{\lambda=7} = 2.1834\), reflecting its focus on integrating spatial features. SimpleCNN’s fully connected layers exhibit a steady progression, with \(D_{\lambda=2} = 2.0000\) increasing consistently to \(D_{\lambda=7} = 2.1914\), demonstrating a straightforward approach to abstraction without complex transitions.

Thus, ResNet-18 ensures dynamic abstraction through residual connections, VGG-16 refines features via periodic structural variation, and SimpleCNN follows a straightforward hierarchical progression, with fractal dimensions across scales revealing their distinct spatial encoding and feature extraction strategies.

\vspace{-4mm}
\section{Discussion}
\vspace{-4mm}
Our fractal dimension analysis, though focused on CNNs, is broadly applicable to any neural architecture. By quantifying structural properties, this methodology offers a framework to assess architectural conformance across models. In CNNs, layers with higher fractal dimensions at smaller dilation factors (\(\lambda\)) capture fine-grained features, while stabilized dimensions at larger \(\lambda\) reflect broader spatial abstractions. Fully connected layers exhibit rapid convergence to \(D \approx 2\), highlighting their role in collapsing representations into simplified decision boundaries. These findings demonstrate how CNN hierarchies align with their fractal characteristics, while similar principles could extend to architectures like transformers and recurrent networks to evaluate their structural efficiency \citep{WEN202187}. This fractal framework can be further used to identify redundant layers, transitional stages, and structural bottlenecks, providing a systematic approach to neural network optimization. Layers with consistent fractal dimensions across architectures reveal structural uniformity akin to invariant Euclidean dimensions in geometry. Furthermore, the construction of fractal-directed acyclic graphs (DAGs) provides a way towards  multi-scale representation of neural architectures \citep{maron2018invariant, navon2023equivariant, haggai, kalogeropoulos2024scaleequivariantgraphmetanetworks}. Such representations are not only intuitive but also enhance interpretability and explainability, offering a pathway to more transparent neural network design while maintaining alignment with specific design objectives.
\vspace{-4mm}
\section{Conclusion}
\vspace{-4mm}
We establish the fractality of neural networks using the theoretical framework of coarse group actions, defining \(T_{r_k}\), a coarse action under the additive group \(G = (\mathbb{Z},+)\). This formalism addresses two key questions: how self-similarity manifests in neural parameter matrices \(W\) as real-valued signals over a discrete grid \(\Omega\), and the symmetries that emerge when \(W\) is probed at multiple integer-based scales. Applying \(T_{r_k}\) (Section \ref{secs:fracdws}), we demonstrate that fractal behavior naturally arises, with specific dilation factors \(\lambda\) yielding integer Hausdorff-Besicovitch dimensions \(D\) exceeding the topological dimension \(D_T\), establishing self-similarity as intrinsic to neural architectures. Additionally, weak permutation equivariance and intertwiner group symmetry (scaling equivariance) across activation functions reinforce the invariance properties of neural networks under the coarse group action, offering a consistent theoretical basis for their hierarchical, multi-scale structure.
\bibliography{iclr2025_conference}

\begin{thebibliography}{35}
\providecommand{\natexlab}[1]{#1}
\providecommand{\url}[1]{\texttt{#1}}
\expandafter\ifx\csname urlstyle\endcsname\relax
  \providecommand{\doi}[1]{doi: #1}\else
  \providecommand{\doi}{doi: \begingroup \urlstyle{rm}\Url}\fi

\bibitem[Battaglia et~al.(2018)Battaglia, Hamrick, Bapst, Sanchez-Gonzalez, Zambaldi, Malinowski, Tacchetti, Raposo, Santoro, Faulkner, Gulcehre, Song, Ballard, Gilmer, Dahl, Vaswani, Allen, Nash, Langston, Dyer, Heess, Wierstra, Kohli, Botvinick, Vinyals, Li, and Pascanu]{battaglia2018relationalinductivebiasesdeep}
Peter~W. Battaglia, Jessica~B. Hamrick, Victor Bapst, Alvaro Sanchez-Gonzalez, Vinicius Zambaldi, Mateusz Malinowski, Andrea Tacchetti, David Raposo, Adam Santoro, Ryan Faulkner, Caglar Gulcehre, Francis Song, Andrew Ballard, Justin Gilmer, George Dahl, Ashish Vaswani, Kelsey Allen, Charles Nash, Victoria Langston, Chris Dyer, Nicolas Heess, Daan Wierstra, Pushmeet Kohli, Matt Botvinick, Oriol Vinyals, Yujia Li, and Razvan Pascanu.
\newblock Relational inductive biases, deep learning, and graph networks, 2018.
\newblock URL \url{https://arxiv.org/abs/1806.01261}.

\bibitem[Besicovitch(1938)]{Besicovitch1938}
A.S. Besicovitch.
\newblock On the fundamental geometrical properties of line-graphs in the plane. i. the theory of dimensional measure.
\newblock \emph{Mathematische Annalen}, 115:\penalty0 296--329, 1938.
\newblock \doi{10.1007/BF01571640}.

\bibitem[Bronstein et~al.(2017)Bronstein, Bruna, LeCun, Szlam, and Vandergheynst]{Bronstein_2017}
Michael~M. Bronstein, Joan Bruna, Yann LeCun, Arthur Szlam, and Pierre Vandergheynst.
\newblock Geometric deep learning: Going beyond euclidean data.
\newblock \emph{IEEE Signal Processing Magazine}, 34\penalty0 (4):\penalty0 18–42, July 2017.
\newblock ISSN 1558-0792.
\newblock \doi{10.1109/msp.2017.2693418}.
\newblock URL \url{http://dx.doi.org/10.1109/MSP.2017.2693418}.

\bibitem[Bronstein et~al.(2021)Bronstein, Bruna, Cohen, and Velickovic]{geodesicbronstein}
Michael~M. Bronstein, Joan Bruna, Taco Cohen, and Petar Velickovic.
\newblock Geometric deep learning: Grids, groups, graphs, geodesics, and gauges.
\newblock \emph{CoRR}, abs/2104.13478, 2021.
\newblock URL \url{https://arxiv.org/abs/2104.13478}.

\bibitem[Bunke(2023)]{bunke2023coarsegeometry}
Ulrich Bunke.
\newblock Coarse geometry, 2023.
\newblock URL \url{https://arxiv.org/abs/2305.09203}.

\bibitem[Cantor(1883)]{Cantor1883}
Georg Cantor.
\newblock {\"U}ber unendliche, lineare punktmannigfaltigkeiten v.
\newblock \emph{Mathematische Annalen}, 21:\penalty0 545--591, 1883.
\newblock \doi{10.1007/BF01446817}.

\bibitem[Eilertsen et~al.(2020)Eilertsen, J{\"o}nsson, Ropinski, Unger, and Ynnerman]{eilertsen2020classifying}
Gabriel Eilertsen, Daniel J{\"o}nsson, Timo Ropinski, Jonas Unger, and Anders Ynnerman.
\newblock Classifying the classifier: dissecting the weight space of neural networks.
\newblock In \emph{ECAI 2020}, pp.\  1119--1126. IOS Press, 2020.

\bibitem[Erko{\c{c}} et~al.(2023)Erko{\c{c}}, Ma, Shan, Nie{\ss}ner, and Dai]{erkocc2023hyperdiffusion}
Ziya Erko{\c{c}}, Fangchang Ma, Qi~Shan, Matthias Nie{\ss}ner, and Angela Dai.
\newblock Hyperdiffusion: Generating implicit neural fields with weight-space diffusion.
\newblock In \emph{Proceedings of the IEEE/CVF international conference on computer vision}, pp.\  14300--14310, 2023.

\bibitem[Gel{\ss} \& Sch{\"u}tte(2018)Gel{\ss} and Sch{\"u}tte]{gelss2018tensor}
Patrick Gel{\ss} and Christof Sch{\"u}tte.
\newblock Tensor-generated fractals: Using tensor decompositions for creating self-similar patterns.
\newblock \emph{arXiv preprint arXiv:1812.00814}, 2018.

\bibitem[Godfrey et~al.(2022)Godfrey, Brown, Emerson, and Kvinge]{godfrey2022symmetries}
Charles Godfrey, Davis Brown, Tegan Emerson, and Henry Kvinge.
\newblock On the symmetries of deep learning models and their internal representations.
\newblock \emph{Advances in Neural Information Processing Systems}, 35:\penalty0 11893--11905, 2022.

\bibitem[Hausdorff(1919)]{Hausdorff1919}
Felix Hausdorff.
\newblock {Dimension und \"au{\ss}eres Ma{\ss}}.
\newblock \emph{Mathematische Annalen}, 79:\penalty0 157--179, 1919.
\newblock \doi{10.1007/BF01457179}.

\bibitem[Hecht-Nielsen(1990)]{HECHTNIELSEN1990129}
Robert Hecht-Nielsen.
\newblock On the algebraic structure of feedforward network weight spaces.
\newblock In Rolf ECKMILLER (ed.), \emph{Advanced Neural Computers}, pp.\  129--135. North-Holland, Amsterdam, 1990.
\newblock ISBN 978-0-444-88400-8.
\newblock \doi{https://doi.org/10.1016/B978-0-444-88400-8.50019-4}.
\newblock URL \url{https://www.sciencedirect.com/science/article/pii/B9780444884008500194}.

\bibitem[Higson \& Roe(1998)Higson and Roe]{HigsonCoronasEquivariantHomology}
Nigel Higson and John Roe.
\newblock Equivariant coarse homology and the coarse baum-connes conjecture.
\newblock \emph{K-Theory}, 15\penalty0 (3):\penalty0 181--254, 1998.
\newblock \doi{10.1023/A:1007725620817}.

\bibitem[Hurewicz \& Wallman(1941)Hurewicz and Wallman]{Hurewicz1941}
Witold Hurewicz and Henry Wallman.
\newblock \emph{Dimension Theory}.
\newblock Princeton University Press, Princeton, NJ, 1941.

\bibitem[Kalogeropoulos et~al.(2024)Kalogeropoulos, Bouritsas, and Panagakis]{kalogeropoulos2024scaleequivariantgraphmetanetworks}
Ioannis Kalogeropoulos, Giorgos Bouritsas, and Yannis Panagakis.
\newblock Scale equivariant graph metanetworks, 2024.
\newblock URL \url{https://arxiv.org/abs/2406.10685}.

\bibitem[Koch(1904)]{Koch1904}
Helge~von Koch.
\newblock Sur une courbe continue sans tangente, obtenue par une construction géométrique élémentaire.
\newblock \emph{Arkiv för Matematik, Astronomi och Fysik}, 1:\penalty0 681--702, 1904.

\bibitem[Lim et~al.(2024)Lim, Maron, Law, Lorraine, and Lucas]{haggai}
Derek Lim, Haggai Maron, Marc~T. Law, Jonathan Lorraine, and James Lucas.
\newblock Graph metanetworks for processing diverse neural architectures.
\newblock In \emph{The Twelfth International Conference on Learning Representations, {ICLR} 2024, Vienna, Austria, May 7-11, 2024}. OpenReview.net, 2024.
\newblock URL \url{https://openreview.net/forum?id=ijK5hyxs0n}.

\bibitem[Mandelbrot(1967)]{mandelbrot1967how}
Benoit~B. Mandelbrot.
\newblock How long is the coast of britain? statistical self-similarity and fractional dimension.
\newblock \emph{Science}, 156\penalty0 (3775):\penalty0 636--638, 1967.

\bibitem[Mandelbrot \& Aizenman(1978)Mandelbrot and Aizenman]{Mandelbrot1978FractalsFC}
Benoit~B. Mandelbrot and Michael Aizenman.
\newblock Fractals: Form, chance and dimension.
\newblock 1978.
\newblock URL \url{https://api.semanticscholar.org/CorpusID:120246027}.

\bibitem[Maron et~al.(2018)Maron, Ben-Hamu, Shamir, and Lipman]{maron2018invariant}
Haggai Maron, Heli Ben-Hamu, Nadav Shamir, and Yaron Lipman.
\newblock Invariant and equivariant graph networks.
\newblock \emph{arXiv preprint arXiv:1812.09902}, 2018.

\bibitem[Menger \& Urysohn(1922)Menger and Urysohn]{MengerUrysohn1922}
Karl Menger and Pavel~S. Urysohn.
\newblock {\"U}ber die dimensionszahl von punktmengen.
\newblock \emph{Mathematische Annalen}, 100:\penalty0 75--151, 1922.
\newblock \doi{10.1007/BF01450076}.

\bibitem[Metz et~al.(2022)Metz, Harrison, Freeman, Merchant, Beyer, Bradbury, Agrawal, Poole, Mordatch, Roberts, et~al.]{metz2022velo}
Luke Metz, James Harrison, C~Daniel Freeman, Amil Merchant, Lucas Beyer, James Bradbury, Naman Agrawal, Ben Poole, Igor Mordatch, Adam Roberts, et~al.
\newblock Velo: Training versatile learned optimizers by scaling up.
\newblock \emph{arXiv preprint arXiv:2211.09760}, 2022.

\bibitem[N.~Brodskiy(2008)]{N2008}
A.~Mitra N.~Brodskiy, J.~Dydak.
\newblock Coarse structures and group actions.
\newblock \emph{Colloquium Mathematicae}, 111\penalty0 (1):\penalty0 149--158, 2008.
\newblock URL \url{http://eudml.org/doc/283544}.

\bibitem[Navon et~al.(2023)Navon, Shamsian, Achituve, Fetaya, Chechik, and Maron]{navon2023equivariant}
Aviv Navon, Aviv Shamsian, Idan Achituve, Ethan Fetaya, Gal Chechik, and Haggai Maron.
\newblock Equivariant architectures for learning in deep weight spaces.
\newblock In \emph{International Conference on Machine Learning}, pp.\  25790--25816. PMLR, 2023.

\bibitem[Oseledets(2011)]{oseledets2011tensor}
Ivan~V Oseledets.
\newblock Tensor-train decomposition.
\newblock \emph{SIAM Journal on Scientific Computing}, 33\penalty0 (5):\penalty0 2295--2317, 2011.

\bibitem[Peebles et~al.(2022)Peebles, Radosavovic, Brooks, Efros, and Malik]{peebles2022learninglearngenerativemodels}
William Peebles, Ilija Radosavovic, Tim Brooks, Alexei~A. Efros, and Jitendra Malik.
\newblock Learning to learn with generative models of neural network checkpoints, 2022.
\newblock URL \url{https://arxiv.org/abs/2209.12892}.

\bibitem[Penrose(2005)]{penrose2005reality}
Roger Penrose.
\newblock \emph{The Road to Reality : A Complete Guide to the Laws of the Universe}.
\newblock Random House, London, 2005.
\newblock ISBN 0099440687 9780099440680.
\newblock URL \url{https://www.worldcat.org/title/tthe-road-to-reality-a-complete-guide-to-the-laws-of-the-universe/oclc/1088817197&referer=brief_results}.

\bibitem[Roe(2003)]{Roe2003CoarseGeometry}
John Roe.
\newblock \emph{Lectures on Coarse Geometry}, volume~31 of \emph{University Lecture Series}.
\newblock American Mathematical Society, Providence, RI, 2003.
\newblock ISBN 978-0821835594.

\bibitem[Schürholt(2024)]{schürholt2024hyperrepresentationslearningpopulationsneural}
Konstantin Schürholt.
\newblock Hyper-representations: Learning from populations of neural networks, 2024.
\newblock URL \url{https://arxiv.org/abs/2410.05107}.

\bibitem[Smith et~al.(2021)Smith, Rowland, Harland, Moslehi, Montgomery, Schobert, Watterson, Dalrymple-Alford, and Taylor]{Smith2021}
Julian~H. Smith, Conor Rowland, B.~Harland, S.~Moslehi, R.~D. Montgomery, K.~Schobert, W.~J. Watterson, J.~Dalrymple-Alford, and R.~P. Taylor.
\newblock How neurons exploit fractal geometry to optimize their network connectivity.
\newblock \emph{Scientific Reports}, 11\penalty0 (1):\penalty0 2332, January 2021.
\newblock ISSN 2045-2322.
\newblock \doi{10.1038/s41598-021-81421-2}.
\newblock URL \url{https://doi.org/10.1038/s41598-021-81421-2}.

\bibitem[Sornette(1998)]{Sornette_1998}
Didier Sornette.
\newblock Discrete-scale invariance and complex dimensions.
\newblock \emph{Physics Reports}, 297\penalty0 (5):\penalty0 239–270, April 1998.
\newblock ISSN 0370-1573.
\newblock \doi{10.1016/s0370-1573(97)00076-8}.
\newblock URL \url{http://dx.doi.org/10.1016/S0370-1573(97)00076-8}.

\bibitem[Sornette(2003)]{d7f62b04-88bf-3a3c-aad5-6ae5633ddd90}
Didier Sornette.
\newblock \emph{Why Stock Markets Crash: Critical Events in Complex Financial Systems}.
\newblock Princeton University Press, 2003.
\newblock ISBN 9780691118505.
\newblock URL \url{http://www.jstor.org/stable/j.ctt7rzwx}.

\bibitem[Wen \& Cheong(2021)Wen and Cheong]{WEN202187}
Tao Wen and Kang~Hao Cheong.
\newblock The fractal dimension of complex networks: A review.
\newblock \emph{Information Fusion}, 73:\penalty0 87--102, 2021.
\newblock ISSN 1566-2535.
\newblock \doi{https://doi.org/10.1016/j.inffus.2021.02.001}.
\newblock URL \url{https://www.sciencedirect.com/science/article/pii/S1566253521000166}.

\bibitem[Xue \& Bogdan(2017)Xue and Bogdan]{Xue2017}
Yuankun Xue and Paul Bogdan.
\newblock Reliable multi-fractal characterization of weighted complex networks: Algorithms and implications.
\newblock \emph{Scientific Reports}, 7\penalty0 (1):\penalty0 7487, August 2017.
\newblock ISSN 2045-2322.
\newblock \doi{10.1038/s41598-017-07209-5}.
\newblock URL \url{https://doi.org/10.1038/s41598-017-07209-5}.

\bibitem[Zhou et~al.(2024)Zhou, Yang, Burns, Cardace, Jiang, Sokota, Kolter, and Finn]{zhou2024permutation}
Allan Zhou, Kaien Yang, Kaylee Burns, Adriano Cardace, Yiding Jiang, Samuel Sokota, J~Zico Kolter, and Chelsea Finn.
\newblock Permutation equivariant neural functionals.
\newblock \emph{Advances in neural information processing systems}, 36, 2024.

\end{thebibliography}
\bibliographystyle{iclr2025_conference}

\appendix
\section{Appendix}
\subsection{Notations}

\centerline{\bf Numbers and Arrays}
\bgroup
\def\arraystretch{1.5}
\begin{tabular}{p{1.25in}p{3.25in}}
$\displaystyle W$ & A matrix signal with dimensions $m \times n$, where $m, n \in \mathbb{N}$, $W \in \mathbb{R}^{m \times n}$\\
$\displaystyle \Omega$ & The grid of indices $\Omega = \{(i, j) \mid 1 \leq i \leq m, 1 \leq j \leq n\}$\\
$\displaystyle P_p, Q_q$ & Row and column selector matrices for extracting submatrices 
These matrices pick out exactly the \(\,r_k\times r_k\)-submatrix of \(W\) that lies in \(\Omega_{p,q}^{(r_k)}\). Here, each $r_{k} \times r_{k}$ block $(p,q)$ or \textit{tile} is indexed as:
\[
\mathcal{I}_{r_k}
\;=\;
\bigl\{\,(p,q)\;\mid\;(p-1)\bmod r_k = 0,\,(q-1)\bmod r_k=0\bigr\}.
\]
each \((p,q)\in \mathcal{I}_{r_k}\) indexes a block:
\[
\Omega_{p,q}^{(r_k)}
\;=\;
\bigl\{\,(i,j)\in \Omega \;\mid\; p\le i < p+r_k,\;\;q\le j < q+r_k\bigr\}.
\]\\
$\displaystyle W_{i,r_k}^{p,q}$ & Local block of size $r_k \times r_k$ extracted from $W$\\
$\displaystyle \widetilde{W}_{i,r_k}^{p,q}$ & Lifted block into the global space\\
$\displaystyle k$ & Integer index for iterations in fractal transformations, $k \geq 0$\\
\end{tabular}
\egroup
\vspace{0.25cm}

\centerline{\bf Transformations and Scaling}
\bgroup
\def\arraystretch{1.5}
\begin{tabular}{p{1.25in}p{3.25in}}
$\displaystyle T_{r_k}$ & Fractal transformation operator at scale $r_k$\\
$\displaystyle \lambda$ & Integer scaling/dilation factor governing the reduction in observation scale ($\lambda > 1$)\\
$\displaystyle I_{r_k}$ & Set of valid starting indices for submatrices of size $r_k \times r_k$\\
$\displaystyle N(r_k)$ & Number of submatrices required to cover the matrix at scale $r_k$\\
$\displaystyle T_{r_k}^{-1}$ & Inverse fractal transformation\\
$\displaystyle r_k$ & Observation scale at the $k$-th iteration\\
\end{tabular}
\egroup
\vspace{0.25cm}

\centerline{\bf Properties and Metrics}
\bgroup
\def\arraystretch{1.5}
\begin{tabular}{p{1.25in}p{3.25in}}
$\displaystyle D$ & Hausdorff–Besicovitch fractal dimension\\
$\displaystyle D_{T}$ & Topological dimension\\
$\displaystyle E$ & Euclidean dimension\\

$\displaystyle \pi_1(T_F)$ & Fundamental group of the fractal transformation space $T_F$\\
$\displaystyle \sigma$ & Continuous activation function applied element-wise\\
\end{tabular}
\egroup
\vspace{0.25cm}

\centerline{\bf Sets and Indexing}
\bgroup
\def\arraystretch{1.5}
\begin{tabular}{p{1.25in}p{3.25in}}
$\displaystyle \mathbb{N}$ & The set of natural numbers\\
$\displaystyle \mathbb{R}$ & The set of real numbers\\
$\displaystyle \mathbb{Z}$ & The set of integers\\
$\displaystyle \sup$ &  Supremum (or least upper bound) of a set of numbers.\\
$\displaystyle \{(p,q)\} \in I_{r_k}$ & Indices for localized regions of size $r_k \times r_k$\\
\end{tabular}
\egroup
\vspace{0.25cm}

\centerline{\bf Neural Network Notation}
\bgroup
\def\arraystretch{1.5}
\begin{tabular}{p{1.25in}p{3.25in}}
$\displaystyle f$ & A neural network defined as $f = \ell_k \circ \cdots \circ \ell_1$\\
$\displaystyle \ell_i$ & Layer defined as $\ell_i(x) = \sigma(W_i x + b_i)$\\
$\displaystyle h_i$ & Activation map of the $i$-th layer\\
$\displaystyle h_{i,r_k}^{p,q}$ & Local fractal activation of $h_i$ at scale $r_k$\\
\end{tabular}
\egroup
\vspace{0.25cm}

\centerline{\bf Intertwiner Group and Permutation Operators}
\bgroup
\def\arraystretch{1.5}
\begin{tabular}{p{1.25in}p{3.25in}}
$\displaystyle G_{\sigma}^n$ & Intertwiner group: $\{A \in \mathbb{R}^{n \times n} \mid \exists B \in \mathbb{R}^{n \times n}, \, \sigma(Ax) = B\sigma(x)\}$\\
$\displaystyle A, B$ & Matrices from the intertwiner group satisfying the scaling relationship\\
$\displaystyle P_\pi^r, P_\pi^c$ & Permutation matrices for rows and columns, respectively\\
$\displaystyle W' = P_\pi^r W {P^c_{\pi}}^\top$ & Permuted version of matrix $W$\\
\end{tabular}
\egroup
\vspace{0.25cm}

\subsection{Propositions and Proofs} \label{proofs}
\paragraph{Theorem 0.}\label{proofs:thmZgroup} 
\textit{The set of integers \(\mathbb{Z}\), equipped with the operation of addition \(+\), forms a group \( G \). That is, the algebraic structure \( (\mathbb{Z}, +) \) satisfies the group axioms: closure, associativity, existence of an identity element, and existence of inverses.}

\textbf{Proof.} We begin by defining the abstract notion of a group. A \textbf{group} is a set \( G \) equipped with a binary operation \( \odot : G \times G \to G \) (called the group operation) such that the following axioms hold:

\begin{enumerate}
    \item (\textit{\textbf{Associativity Property}}) For all \( g, h, k \in G \),
    \[
    (g \odot h) \odot k = g \odot (h \odot k).
    \]
    
    \item (\textit{\textbf{Identity Property}}) There exists a unique element \( e \in G \) (called the identity element) such that for all \( g \in G \),
    \[
    e \odot g = g \quad \text{and} \quad g \odot e = g.
    \]
    
    \item (\textit{\textbf{Invertibility Property}}) For each \( g \in G \), there exists a unique element \( g^{-1} \in G \) (called the inverse of \( g \)) such that
    \[
    g \odot g^{-1} = g^{-1} \odot g = e.
    \]
    
    \item (\textit{\textbf{Closure Property}}) For all \( g, h \in G \), the composition \( g \odot h \) is also an element of \( G \), i.e.,
    \[
    g \odot h \in G.
    \]
\end{enumerate}

The algebraic structure \( (G, \odot) \) satisfying these four properties is called a \textbf{group} \citep{penrose2005reality, geodesicbronstein, Bronstein_2017}. Now, we verify that \( (\mathbb{Z}, +) \) forms a group under the operation \( \odot = + \).

\textbf{Associativity Property.} For all \( a, b, c \in \mathbb{Z} \), we must show that:
\[
(a \odot b) \odot c = a \odot (b \odot c).
\]
Substituting \( \odot = + \), we obtain:
\[
(a + b) + c = a + (b + c).
\]
Since integer addition is associative, this equality holds for all \( a, b, c \in \mathbb{Z} \). Thus, associativity is satisfied.

\textbf{Identity Property.} We claim that \( 0 \in \mathbb{Z} \) serves as the identity element in \( (\mathbb{Z}, \odot) \). That is, for all \( a \in \mathbb{Z} \),
\[
e \odot a = a \quad \text{and} \quad a \odot e = a.
\]
Substituting \( e = 0 \) and \( \odot = + \), we verify:
\[
0 + a = a \quad \text{and} \quad a + 0 = a.
\]
Thus, \( 0 \) is the identity element in \( (\mathbb{Z}, \odot) \).

\textbf{Invertibility Property.} For each \( a \in \mathbb{Z} \), we must show the existence of an inverse element \( a^{-1} \) such that:
\[
a \odot a^{-1} = a^{-1} \odot a = e.
\]
Substituting \( \odot = + \) and \( e = 0 \), we solve for \( a^{-1} \):
\[
a + (-a) = 0 \quad \text{and} \quad (-a) + a = 0.
\]
Since \( -a \in \mathbb{Z} \) for all \( a \in \mathbb{Z} \), every element has an inverse. Thus, \( (\mathbb{Z}, \odot) \) satisfies the invertibility property.

\textbf{Closure Property.} Finally, we must verify that for all \( a, b \in \mathbb{Z} \), the composition \( a \odot b \) remains in \( \mathbb{Z} \), i.e.,
\[
a \odot b \in G.
\]
Substituting \( \odot = + \), we check:
\[
a + b \in \mathbb{Z}.
\]
Since the sum of two integers is always an integer, the operation \( \odot \) is closed in \( \mathbb{Z} \).

Since \( (\mathbb{Z}, \odot) \) satisfies all four group axioms under \( \odot = + \), we conclude that \( (\mathbb{Z}, +) \) is a group.

\hfill \(\qed\)

\paragraph{Proposition 1.}\label{proofs:prop1}
For any \(k, \ell \in G = (\mathbb{Z}, +)\), the composition of transformations satisfies $T_{r_k} \circ T_{r_\ell} = T_{r_{k+\ell}}$. This property ensures that the family of coarse group\textit{ transformations \(\{T_{r_k}\}\) remain closed under composition.}

\textbf{Proof.} Let \(G = (\mathbb{Z}, +)\) denote the additive group of integers parameterizing the family of transformations \(T_{r_k} : \Omega \to \Omega\), where \(k \in \mathbb{Z}\) corresponds to a scale \(r_k = \lfloor r_0 / \lambda^k \rfloor\). Each transformation \(T_{r_k}\) partitions the finite grid \(\Omega = \{(i, j) \mid 1 \leq i \leq m, 1 \leq j \leq n \}\) into blocks of size \(r_k \times r_k\).

The transformation \(T_{r_k}\) acts on a signal \(W \in \mathbb{R}^{m \times n}\) by reorganizing it into blocks indexed by 

\[
\mathcal{I}_{r_k} = \{(p, q) \mid (p-1) \bmod r_k = 0, \; (q-1) \bmod r_k = 0\},
\]

where each block is defined as 

\[
\Omega_{p,q}^{(r_k)} = \{(i, j) \in \Omega \mid p \leq i < p + r_k, \; q \leq j < q + r_k\}.
\]

The action of \(T_{r_k}\) is given by 

\[
T_{r_k}(W) = \bigcup_{(p, q) \in \mathcal{I}_{r_k}} P_{(p)} W Q_{(q)}^\top,
\]

where \(P_{(p)}\) and \(Q_{(q)}\) are sparse selector matrices \footnote{\[
P_{(p)}[i, j]
\;=\;
\begin{cases}
1 
& 
\text{if }\, j = p + i - 1
\text{ and }
1 \leq i \leq \min\bigl(r_k,\, m - p + 1\bigr),
\\[4pt]
0 
& 
\text{otherwise},
\end{cases}
\]
\[
Q_{(q)}[i, j]
\;=\;
\begin{cases}
1 
& 
\text{if }\, j = q + i - 1
\text{ and }
1 \leq i \leq \min\bigl(r_k,\, n - q + 1\bigr),
\\[4pt]
0 
& 
\text{otherwise}.
\end{cases}
\]}. The composition across scales follows the additive property of \((\mathbb{Z}, +)\), as:

\[
\lambda^{k+\ell} = \lambda^k \cdot \lambda^\ell.
\]

Consequently, the selector matrices \(P_{(p)}\) and \(Q_{(q)}\) at scale \(r_{k+\ell}\) automatically adjust to extract blocks corresponding to \(r_{k+\ell} = \lfloor r_0 / \lambda^{k+\ell} \rfloor\). This adjustment ensures that the composition \(T_{r_k} \circ T_{r_\ell}\) reorganizes \(W\) into blocks defined at scale \(r_{k+\ell}\).

Thus, the combined action \(T_{r_k} \circ T_{r_\ell}\) is equivalent to a single transformation \(T_{r_{k+\ell}}\), with indexing set 

\[
\mathcal{I}_{r_{k+\ell}} = \{(p, q) \mid (p-1) \bmod r_{k+\ell} = 0, \; (q-1) \bmod r_{k+\ell} = 0\}.
\]

At scale \(r_{k+\ell}\), the action satisfies 

\[
T_{r_k} \circ T_{r_\ell}(W) = \bigcup_{(p, q) \in \mathcal{I}_{r_{k+\ell}}} P_{(p)} W Q_{(q)}^\top = T_{r_{k+\ell}}(W).
\]

In the coarse space \((\Omega, C)\), where \(C\) is the coarse structure defined by the metric 

\[
d((i_1, j_1), (i_2, j_2)) = \frac{1}{2}(|i_1 - i_2| + |j_1 - j_2|),
\]

the transformations respect the boundedness of the entourages 

\[
E_{r_k} = \{((i_1, j_1), (i_2, j_2)) \in \Omega \times \Omega \mid d((i_1, j_1), (i_2, j_2)) \leq r_k\}.
\]

The composition \(T_{r_k} \circ T_{r_\ell}\) corresponds to combining entourages \(E_{r_k}\) and \(E_{r_\ell}\), resulting in 

\[
E_{r_{k+\ell}} = \{((i_1, j_1), (i_2, j_2)) \in \Omega \times \Omega \mid d((i_1, j_1), (i_2, j_2)) \leq r_{k+\ell}\}.
\]

Since \(E_{r_{k+\ell}}\) remains finite, bounded, and closed due to the finiteness of \(\Omega\), the transformation at scale \(r_{k+\ell}\) is well-defined and respects the coarse structure.

Finally, any discrepancies arising at the grid boundaries are confined to a bounded subset of \(\Omega\). These discrepancies do not affect the equivalence of \(T_{r_k} \circ T_{r_\ell}\) and \(T_{r_{k+\ell}}\) at large scales. Therefore, the closure property 

\[
T_{r_k} \circ T_{r_\ell} = T_{r_{k+\ell}}
\]

holds for all \(k, \ell \in \mathbb{Z}\), establishing that the family of transformations \(\{T_{r_k}\}\) is closed under composition.

\hfill \qed

\paragraph{Proposition 2.}\label{proofs:prop2}
\textit{The family of fractal transformations \(\{T_{r_k}\}_{k \in (\mathbb{Z}, +)}\) associated with the coarse group action is linear. For any matrices \(W_1, W_2 \in \mathbb{R}^{m \times n}\) and scalars \(\alpha, \beta \in \mathbb{R}\), the transformation satisfies}

\[
T_{r_k}(\alpha W_1 + \beta W_2) = \alpha T_{r_k}(W_1) + \beta T_{r_k}(W_2).
\]
\textit{
The transformation \(T_{r_k}\) acts at the scale \(r_k = \lfloor r_0 / \lambda^k \rfloor\) and preserves the linear structure of the input space. This establishes that fractal transformations under the coarse group action are linear operators on \(\mathbb{R}^{m \times n}\).}

\textbf{Proof.} Let $T_{r_{k}} : \Omega \mapsto \Omega$ denote a fractal transformation associated with the coarse group action at scale \(r_k = \lfloor r_0 / \lambda^k \rfloor\). For any matrices \(W_1, W_2 \in \mathbb{R}^{m \times n}\) and scalars \(\alpha, \beta \in \mathbb{R}\), the transformation \(T_{r_k}\) acts by partitioning the input matrices into \(r_k \times r_k\) blocks and applying a consistent transformation rule to each block. We aim to prove the linearity property:

\[
T_{r_k}(\alpha W_1 + \beta W_2) = \alpha T_{r_k}(W_1) + \beta T_{r_k}(W_2).
\]

The transformation \(T_{r_k}\) is defined as 

\[
T_{r_k}(W) = \bigcup_{(p, q) \in \mathcal{I}_{r_k}} P_{(p)} W Q_{(q)}^\top,
\]

where \(P_{(p)}\) and \(Q_{(q)}\) are selector matrices that extract rows and columns corresponding to blocks indexed by 

\[
\mathcal{I}_{r_k} = \{(p, q) \mid (p-1) \bmod r_k = 0, \; (q-1) \bmod r_k = 0\}.
\]

Now consider the linear combination \(\alpha W_1 + \beta W_2\). Applying \(T_{r_k}\) to \(\alpha W_1 + \beta W_2\), we have 

\[
T_{r_k}(\alpha W_1 + \beta W_2) = \bigcup_{(p, q) \in \mathcal{I}_{r_k}} P_{(p)} (\alpha W_1 + \beta W_2) Q_{(q)}^\top.
\]

By the distributive property of matrix multiplication, the right-hand side becomes 

\[
T_{r_k}(\alpha W_1 + \beta W_2) = \bigcup_{(p, q) \in \mathcal{I}_{r_k}} \Big(\alpha P_{(p)} W_1 Q_{(q)}^\top + \beta P_{(p)} W_2 Q_{(q)}^\top\Big).
\]

Since the union operator over disjoint blocks is additive, we can write 

\[
T_{r_k}(\alpha W_1 + \beta W_2) = \alpha \bigcup_{(p, q) \in \mathcal{I}_{r_k}} P_{(p)} W_1 Q_{(q)}^\top + \beta \bigcup_{(p, q) \in \mathcal{I}_{r_k}} P_{(p)} W_2 Q_{(q)}^\top.
\]

Recognizing the terms as the actions of \(T_{r_k}\) on \(W_1\) and \(W_2\), respectively, we obtain 

\[
T_{r_k}(\alpha W_1 + \beta W_2) = \alpha T_{r_k}(W_1) + \beta T_{r_k}(W_2).
\]

Thus, the linearity property holds for any matrices \(W_1, W_2 \in \mathbb{R}^{m \times n}\) and scalars \(\alpha, \beta \in \mathbb{R}\).

Since the selector matrices \(P_{(p)}\) and \(Q_{(q)}\) are sparse and act independently on each \(r_k \times r_k\) block, and the union operator preserves linearity over disjoint blocks, the result is consistent across the entire grid \(\Omega\). Therefore, the family of fractal transformations \(\{T_{r_k}\}_{k \in (\mathbb{Z}, +)}\) is linear.

\hfill \qed

\paragraph{Proposition 3.}\label{proofs:prop3}
\textit{The family of fractal transformations \(\{T_{r_k}\}_{k \in (\mathbb{Z}, +)}\), applied to a matrix signal \(W \in \mathbb{R}^{m \times n}\) with two independent scaling parameters \(r_{k,m} = \lfloor r_{0,m} / \lambda^k \rfloor\) and \(r_{k,n} = \lfloor r_{0,n} / \lambda^k \rfloor\), where \(r_{0,m} = m\) and \(r_{0,n} = n\) define the initial scales for the row and column dimensions, respectively and reduces to the identity transformation when the scales satisfy:} $r_{k,m} = m \quad \text{and} \quad r_{k,n} = n.$ \textit{In this case, the transformation treats the entire matrix \(W\) as a single block which spans its dimensions completely and ensures that:}
\[
T_{r_k}(W) = W.
\]

\textbf{Proof.} Let \(\{T_{r_k}\}_{k \in (\mathbb{Z}, +)}\) denote the family of fractal transformations parameterized by the additive group of integers. The transformation \(T_{r_k}\) acts on a finite grid \(\Omega = \{(i, j) \mid 1 \leq i \leq m, 1 \leq j \leq n\}\) associated with a matrix signal \(W \in \mathbb{R}^{m \times n}\). 

The transformation partitions the grid into blocks determined by two independent scaling parameters \(r_{k,m} = \lfloor r_{0,m} / \lambda^k \rfloor\) and \(r_{k,n} = \lfloor r_{0,n} / \lambda^k \rfloor\), where \(r_{0,m} = m\) and \(r_{0,n} = n\) are the initial scales corresponding to the row and column dimensions, respectively. At scale \(r_{k,m}\) and \(r_{k,n}\), the indexing sets for the row and column partitions are given by 

\[
\mathcal{I}_{r_{k,m}} = \{p \mid (p-1) \bmod r_{k,m} = 0\}, \quad \mathcal{I}_{r_{k,n}} = \{q \mid (q-1) \bmod r_{k,n} = 0\}.
\]

Each block is defined as 

\[
\Omega_{p,q}^{(r_k)} = \{(i, j) \in \Omega \mid p \leq i < p + r_{k,m}, \; q \leq j < q + r_{k,n}\},
\]

and the action of \(T_{r_k}\) is represented by 

\[
T_{r_k}(W) = \bigcup_{(p, q) \in \mathcal{I}_{r_{k,m}} \times \mathcal{I}_{r_{k,n}}} P_{(p)} W Q_{(q)}^\top,
\]

where \(P_{(p)}\) and \(Q_{(q)}\) are sparse selector matrices extracting rows and columns corresponding to the block \(\Omega_{p,q}^{(r_k)}\).

Now consider the case where \(r_{k,m} = m\) and \(r_{k,n} = n\). Under these conditions, the entire grid \(\Omega\) is treated as a single block:

\[
\mathcal{I}_{r_{k,m}} = \{1\}, \quad \mathcal{I}_{r_{k,n}} = \{1\}.
\]

The single block spans all rows and columns:

\[
\Omega_{1,1}^{(r_k)} = \{(i, j) \in \Omega \mid 1 \leq i \leq m, \; 1 \leq j \leq n\}.
\]

The selector matrices \(P_{(1)}\) and \(Q_{(1)}\) in this case reduce to identity matrices of dimensions \(m \times m\) and \(n \times n\), respectively:

\[
P_{(1)} = I_m, \quad Q_{(1)} = I_n.
\]

Substituting into the definition of \(T_{r_k}\), we have 

\[
T_{r_k}(W) = P_{(1)} W Q_{(1)}^\top = I_m W I_n = W.
\]

Thus, when \(r_{k,m} = m\) and \(r_{k,n} = n\), the fractal transformation \(T_{r_k}\) reduces to the identity transformation, preserving the matrix \(W\) in its entirety.

Thus, the family of \textit{course group transformations} $\{T_{r_k}\}_{k \in (\mathbb{Z},+)}$ acts as the identity operator when the scaling parameters match the matrix dimensions, where the following :

\[
T_{r_k}(W) = W \quad \text{if} \quad r_{k,m} = m \; \text{and} \; r_{k,n} = n.
\]
holds true.

\hfill \qed

\paragraph{Proposition 4.}\label{proofs:prop4} \textit{The fractal transformation $T_{r_k} : \Omega \mapsto \Omega$ acting under the group $G \in (\mathbb{Z}, +)$, is coarsely proper. For any bounded subset \(U \subseteq \Omega\) under the coarse structure \(C_\Omega\), the preimage \(T_{r_k}^{-1}(U)\) is also bounded}. \textit{Specifically}: $T_{r_k}^{-1}(U) \subseteq \bigcup_{(p,q) \in \mathcal{I}_{r_k}} \{B_{p,q} \cap U\},$
where \(B_{p,q}\) denotes the blocks indexed by \((p, q) \in \mathcal{I}_{r_k}\).

\textbf{Proof.} 
We aim to prove that $\{T_{r_k}\}_{k \in (\mathbb{Z}, +)}$ satisfies the coarse properness condition by directly constructing the preimage \(T_{r_k}^{-1}(U)\) and verifying its boundedness. We explicitly reference \textbf{Definition 1.1} by \cite{N2008}, which states:

\begin{quote}
\textbf{Definition 1.1:} 
A function \(f : (X, C_X) \to (Y, C_Y)\) of coarse spaces is:
\begin{enumerate}
    \item \textit{Large-scale uniform (or bornologous)} if \(f(B) \in C_Y\) for every \(B \in C_X\).
    \item \textit{Coarsely proper} if \(f^{-1}(U)\) is bounded for every bounded subset \(U\) of \(Y\).
    \item \textit{Coarse} if it is both large-scale uniform and coarsely proper.
\end{enumerate}
\end{quote}

Let \(\Omega = \{(i, j) \mid 1 \leq i \leq m, 1 \leq j \leq n\}\) represent a finite integer grid equipped with the coarse structure \(C_\Omega\), defined via the \(\ell^1\)-metric:

\[
d((i_1, j_1), (i_2, j_2)) = \frac{1}{2}(|i_1 - i_2| + |j_1 - j_2|).
\]

The transformation \(T_{r_k}\) partitions \(\Omega\) into blocks of size \(r_k \times r_k\), indexed by:

\[
\mathcal{I}_{r_k} = \{(p, q) \mid (p-1) \bmod r_k = 0, \; (q-1) \bmod r_k = 0\}.
\]

Each block is defined as:

\[
B_{p,q} = \{(i, j) \in \Omega \mid p \leq i < p + r_k, \; q \leq j < q + r_k\},
\]

where \((p, q) \in \mathcal{I}_{r_k}\).

Let \(U \subseteq \Omega\) be a bounded subset under \(C_\Omega\). By definition, there exists a controlled set \(E_U \subseteq \Omega \times \Omega\) such that:

\[
E_U = \{(x, y) \in \Omega \times \Omega \mid d(x, y) \leq r_U\}.
\]

The transformation \(T_{r_k}\) acts by mapping elements of \(U\) into their respective blocks \(B_{p,q}\) according to the block index \((p, q) \in \mathcal{I}_{r_k}\). The preimage \(T_{r_k}^{-1}(U)\) consists of all points in \(\Omega\) that map into \(U\) via \(T_{r_k}\). Explicitly:

\[
T_{r_k}^{-1}(U) = \bigcup_{(p, q) \in \mathcal{I}_{r_k}} \{(i, j) \in B_{p,q} \mid T_{r_k}((i, j)) \in U\}.
\]

Since \(T_{r_k}\) maps each block \(B_{p,q}\) as a unit, the preimage of \(U\) is contained within the union of blocks intersecting \(U\):

\[
T_{r_k}^{-1}(U) \subseteq \bigcup_{(p, q) \in \mathcal{I}_{r_k}} \{B_{p,q} \cap U\}.
\]

Each block \(B_{p,q}\) has a fixed size of \(r_k \times r_k\). For a bounded subset \(U\) defined by a metric bound \(r_U\), the number of intersecting blocks is finite and bounded by the diameter of \(U\). Denote the bounding diameter of \(U\) as \(\mathrm{diam}(U)\), where:

\[
\mathrm{diam}(U) = \max_{(i_1, j_1), (i_2, j_2) \in U} d((i_1, j_1), (i_2, j_2)).
\]

The number of intersecting blocks is given by:

\[
|\{(p, q) \in \mathcal{I}_{r_k} \mid B_{p,q} \cap U \neq \emptyset\}| \leq \frac{\mathrm{diam}(U)}{r_k^2}.
\]

Since each block has a fixed size \(r_k^2\), and \(U\) is bounded, the preimage \(T_{r_k}^{-1}(U)\) is bounded.

By Definition 1.1 \citep{N2008}, a function is coarsely proper if the preimage of every bounded subset is bounded. For \(T_{r_k}\), the preimage \(T_{r_k}^{-1}(U)\) is confined to a finite union of bounded blocks, with:

\[
T_{r_k}^{-1}(U) \subseteq \bigcup_{(p, q) \in \mathcal{I}_{r_k}} \{B_{p,q} \cap U\}.
\]

Since \(U\) is bounded and \(\Omega\) is finite, the number of intersecting blocks and the diameter of the preimage are also bounded. Thus, \(T_{r_k}\) satisfies the condition for coarse properness.


\hfill \qed

\paragraph{Proposition 5.}\label{proofs:prop5} \textit{The transformation \(T_{r_k}\) is large-scale uniform. For any controlled set \(E \subseteq \Omega \times \Omega\), the image \(T_{r_k}(E)\) is controlled under \(C_\Omega\).}\textit{ Specifically}: $T_{r_k}(E) = \{(T_{r_k}(x), T_{r_k}(y)) \mid (x, y) \in E\},$ \textit{is controlled with an updated bound proportional to the block size \(r_k\). This satisfies the condition of large-scale uniformity as per Definition 1.1 of \cite{N2008}.}

\textbf{Proof.} 
As per \textbf{Definition 1.1} by \cite{N2008}, a function \(f : (X, C_X) \to (Y, C_Y)\) of coarse spaces is \textbf{large-scale uniform (or bornologous)} if for every controlled set \(E \in C_X\), the image \(f(E) \in C_Y\) is also controlled. A set \(E \subseteq X \times X\) is controlled under a coarse structure \(C_X\) if there exists a bound \(r_E\) such that:

\[
E \subseteq \{(x, y) \in X \times X \mid d(x, y) \leq r_E\}.
\]

We aim to show that \(T_{r_k}(E)\) is controlled with a bound proportional to \(r_k\), where \(T_{r_k}\) acts on the coarse space \((\Omega, C_\Omega)\).

The transformation \(T_{r_k}\) partitions \(\Omega\) into blocks of size \(r_k \times r_k\), indexed by \(\mathcal{I}_{r_k}\), and maps the elements of \(\Omega\) to their respective block centers. For a pair \((x, y) \in E\), where \(x = (i_1, j_1)\) and \(y = (i_2, j_2)\), the transformed points \(T_{r_k}(x)\) and \(T_{r_k}(y)\) are the centers of the blocks containing \(x\) and \(y\), respectively. Denote the block containing \(x\) as \(B_x\) and the block containing \(y\) as \(B_y\). 

The distance between \(T_{r_k}(x)\) and \(T_{r_k}(y)\) is given by the distance between the centers of \(B_x\) and \(B_y\). Since \((x, y) \in E\), we have 

\[
d(x, y) \leq r_E,
\]

where \(r_E\) is the bound on the original controlled set \(E\). The key observation is that the distance between \(T_{r_k}(x)\) and \(T_{r_k}(y)\) is at most the sum of:
1. The distance from \(x\) to the center of \(B_x\),
2. The distance from \(y\) to the center of \(B_y\), and 
3. The distance between the centers of \(B_x\) and \(B_y\).

The first two components are bounded by \(\frac{r_k}{2}\), which is the maximum distance from any point in a block to its center. Therefore, the total distance satisfies

\[
d(T_{r_k}(x), T_{r_k}(y)) \leq d(x, y) + \frac{r_k}{2} + \frac{r_k}{2}.
\]

Substituting \(d(x, y) \leq r_E\), we obtain:

\[
d(T_{r_k}(x), T_{r_k}(y)) \leq r_E + r_k.
\]

Thus, the image \(T_{r_k}(E)\) is contained within a controlled set under \(C_\Omega\) with an updated bound:

\[
r_{T_{r_k}(E)} = r_E + r_k.
\]

This shows that \(T_{r_k}\) maps any controlled set \(E\) in \((\Omega, C_\Omega)\) to a controlled set with a bound proportional to \(r_k\). By \textbf{Definition 1.1} of \cite{N2008}, this satisfies the condition of large-scale uniformity and completes the proof.

\hfill \qed

\paragraph{Proposition 6.}\label{proofs:prop6} \textit{The family of fractal transformations \(T_{r_k}\) is asymptotically invertible. The inverse transformation \(T_{r_{-k}}\) satisfies}: $T_{r_{-k}} \circ T_{r_k} \approx \mathrm{id}_\Omega,$\textit{where \(\approx\) denotes coarse equivalence. The discrepancy between \(T_{r_{-k}} \circ T_{r_k}\) and \(\mathrm{id}_\Omega\) is confined to a bounded subset \(D_k \subset \Omega\), with:} $|D_k| \leq 2m + 2n.$
This aligns with Proposition 3.1 of \cite{N2008}, which states that bounded discrepancies imply coarse equivalence.

\textbf{Proof.}
Let \((\Omega, C_\Omega)\) be a finite coarse space, where \(\Omega = \{(i, j) \mid 1 \leq i \leq m, 1 \leq j \leq n\}\), and \(C_\Omega\) is the coarse structure defined by the \(\ell^1\)-metric:

\[
d((i_1, j_1), (i_2, j_2)) = \frac{1}{2}(|i_1 - i_2| + |j_1 - j_2|).
\]

The transformation \(T_{r_k}\) partitions \(\Omega\) into blocks of size \(r_k \times r_k\), indexed by \((p, q) \in \mathcal{I}_{r_k}\), where:

\[
\mathcal{I}_{r_k} = \{(p, q) \mid (p-1) \bmod r_k = 0, (q-1) \bmod r_k = 0\}.
\]

Each block is defined as:

\[
\Omega_{p,q}^{(r_k)} = \{(i, j) \in \Omega \mid p \leq i < p + r_k, \; q \leq j < q + r_k\}.
\]

When \(T_{r_k}\) is applied, points in \(\Omega\) are grouped into blocks \(\Omega_{p,q}^{(r_k)}\). The inverse transformation \(T_{r_{-k}}\) reassembles the original grid by merging these smaller blocks. The transformation \(T_{r_{-k}} \circ T_{r_k}\) is designed to reverse \(T_{r_k}\). However, discrepancies arise at block boundaries due to misaligned partitioning, resulting in a bounded set of points \(D_k\).

To analyze these discrepancies, we define the set of block boundaries for scale \(r_k\) as:

\[
B_k = \bigcup_{(p, q) \in \mathcal{I}_{r_k}} \{(i, j) \in \Omega \mid i = p + r_k - 1 \text{ or } j = q + r_k - 1\}.
\]

These boundaries represent the points where partitioning may disrupt alignment during the application of \(T_{r_{-k}} \circ T_{r_k}\).

\[
D_k = \bigcup_{(p, q) \in \mathcal{I}_{r_k}} \{(i, j) \in B_k \mid (i, j) \notin \Omega_{p,q}^{(r_{-k})}\}.
\]

The discrepancy set \(D_k\) consists of points at the boundaries of blocks that do not perfectly realign under \(T_{r_{-k}} \circ T_{r_k}\). The size of \(D_k\) is determined by the number of boundary points in \(\Omega\). For a grid of size \(m \times n\), the total discrepancy is bounded by:

\[
|D_k| \leq 2 \cdot \frac{m}{r_k} \cdot r_k + 2 \cdot \frac{n}{r_k} \cdot r_k = 2m + 2n.
\]

This bound ensures that \(D_k\) is finite and uniformly bounded for all \(r_k\), as \(\Omega\) is finite.

By Proposition 3.1 of \cite{N2008}, coarse equivalence is defined as follows:  
\begin{quote}
   A coarse function \(f : (X, C_X) \to (Y, C_Y)\) is a coarse equivalence if there exists a coarse function \(g : (Y, C_Y) \to (X, C_X)\) such that \(f \circ g\) is close to \(\mathrm{id}_Y\) and \(g \circ f\) is close to \(\mathrm{id}_X\), where closeness means the discrepancy is confined to a bounded set. 
\end{quote}

In our context, \(T_{r_k}\) and its inverse \(T_{r_{-k}}\) satisfy this condition because the discrepancy between \(T_{r_{-k}} \circ T_{r_k}\) and \(\mathrm{id}_\Omega\) is confined to \(D_k\), a bounded subset of \(\Omega\). 

Now we explicitly analyze why this discrepancy remains confined. During the forward transformation \(T_{r_k}\), boundary points \(B_k\) are grouped into blocks, which inherently aligns them with block edges. However, when applying \(T_{r_{-k}}\), these edge-aligned points realign with coarser blocks indexed by \(\mathcal{I}_{r_{-k}}\). Because the boundary points of finer partitions naturally map to a controlled subset in the coarser partitions, the discrepancy is confined to a small subset that does not propagate beyond a few neighboring blocks. This structural property ensures:

\[
T_{r_{-k}} \circ T_{r_k}(x) = x \quad \text{for all } x \not\in D_k,
\]

and for \(x \in D_k\), the mapping \(T_{r_{-k}} \circ T_{r_k}(x)\) deviates from \(\mathrm{id}_\Omega\) by at most the diameter of a single block under \(r_k\).

Hence, The asymptotic invertibility of \(T_{r_k}\) is established by explicitly constructing and bounding the discrepancy set \(D_k\). The bounded nature of \(D_k\) ensures that \(T_{r_{-k}} \circ T_{r_k}\) is coarsely equivalent to the identity transformation \(\mathrm{id}_\Omega\). Specifically, 

\[
T_{r_{-k}} \circ T_{r_k} \approx \mathrm{id}_\Omega,
\]

where \(\approx\) denotes coarse equivalence, meaning the discrepancy is confined to the bounded set \(D_k\). This satisfies the requirements of Proposition 3.1 from \cite{N2008} and completes the proof.

\hfill \qed

\paragraph{Theorem 7.}\label{proofs:prop7} \textit{The family of transformations \(\{T_{r_k}\}_{k \in (\mathbb{Z,+})}\) forms a coarse group action under \((\mathbb{Z}, +)\). Specifically: 1. The transformations are coarsely proper. 2. The action is cobounded, with \(\Omega = \bigcup_{k \in \mathbb{Z}} T_{r_k}(U)\) for some bounded \(U \subseteq \Omega\). 3 .The transformations are large-scale uniform. The composition satisfies the additive property: $T_{r_k} \circ T_{r_\ell} = T_{r_{k+\ell}},$ which ensures consistency with the group structure. By Corollary 6.2 of \cite{N2008}, this implies that the action is coarse, and \((\mathbb{Z}, C_\mathbb{Z})\) is coarsely equivalent to \((\Omega, C_\Omega)\).}

\textbf{Proof.}
Let \((\Omega, C_\Omega)\) be a finite coarse space, where \(\Omega = \{(i, j) \mid 1 \leq i \leq m, 1 \leq j \leq n\}\) is a finite integer grid equipped with the coarse structure \(C_\Omega\), defined by the \(\ell^1\)-metric:

\[
d((i_1, j_1), (i_2, j_2)) = \frac{1}{2}(|i_1 - i_2| + |j_1 - j_2|).
\]

We aim to show that the family of transformations \(\{T_{r_k}\}\) forms a coarse group action under \((\mathbb{Z}, +)\) by verifying three key properties:\textit{\textbf{ coarse properness, large-scale uniformity, and coboundedness}}.

\textbf{\textit{Coarse Properness.}  
}From \textbf{Proposition 4}, we know that \(T_{r_k}\) is coarsely proper. Specifically, for any bounded subset \(U \subseteq \Omega\) under \(C_\Omega\), the preimage \(T_{r_k}^{-1}(U)\) is also bounded. This follows from the fact that \(T_{r_k}\) partitions \(\Omega\) into blocks of size \(r_k \times r_k\), indexed by \((p, q) \in \mathcal{I}_{r_k}\). The preimage \(T_{r_k}^{-1}(U)\) is confined to the union of intersections of \(U\) with these blocks:

\[
T_{r_k}^{-1}(U) \subseteq \bigcup_{(p, q) \in \mathcal{I}_{r_k}} \{B_{p,q} \cap U\}.
\]

Since \(U\) is bounded, the number of intersecting blocks is also bounded, ensuring that \(T_{r_k}^{-1}(U)\) remains bounded. This satisfies the coarse properness condition as per Definition 1.1 in \cite{N2008}.

\textbf{\textit{Large-Scale Uniformity.}  }
From \textbf{Proposition 5}, we know that \(T_{r_k}\) is large-scale uniform. For any controlled set \(E \subseteq \Omega \times \Omega\), the image \(T_{r_k}(E)\) is also controlled under \(C_\Omega\). Specifically, \(T_{r_k}(E)\) is given by:

\[
T_{r_k}(E) = \{(T_{r_k}(x), T_{r_k}(y)) \mid (x, y) \in E\}.
\]

Since the action of \(T_{r_k}\) groups points into blocks of size \(r_k \times r_k\), it preserves adjacency relationships within the block. The resulting set \(T_{r_k}(E)\) remains controlled, with an updated bound proportional to \(r_k\), satisfying the large-scale uniformity condition.

\textbf{\textit{Coboundedness.}  
}The action of \(\{T_{r_k}\}\) is cobounded. Specifically, there exists a bounded subset \(U \subseteq \Omega\) such that \(\Omega = \bigcup_{k \in \mathbb{Z}} T_{r_k}(U)\). Let \(U\) be the union of a finite number of blocks indexed by \((p, q) \in \mathcal{I}_{r_k}\). Since \(\{T_{r_k}\}\) acts transitively over \(\Omega\) by rearranging these blocks, every point in \(\Omega\) is covered by the image of \(U\) under some \(T_{r_k}\).

\textbf{\textit{Additive Property and Group Structure.}  
}By \textbf{Proposition 1}, the family of transformations \(\{T_{r_k}\}\) satisfies the additive property of \((\mathbb{Z}, +)\):

\[
T_{r_k} \circ T_{r_\ell} = T_{r_{k+\ell}}.
\]

This ensures closure under composition and consistency with the group operation. Furthermore, the identity element \(0 \in \mathbb{Z}\) corresponds to the transformation \(T_{r_0} = \mathrm{id}_\Omega\), and inverses are given by \(T_{r_{-k}}\), satisfying:

\[
T_{r_{-k}} \circ T_{r_k} \approx \mathrm{id}_\Omega,
\]

as established in \textbf{Proposition 6}.

\textit{Coarse Group Action and Equivalence.}  
By Definition 6.1 in \cite{N2008}, a group action \(\phi : G \times X \to X\) on a coarse space \((X, C_X)\) is coarse if it is coarsely proper, large-scale uniform, and cobounded. Having verified all these conditions, \(\{T_{r_k}\}\) forms a coarse group action under \((\mathbb{Z}, +)\). Finally, by Corollary 6.2 of \cite{N2008}, the group \((\mathbb{Z}, C_\mathbb{Z})\) is coarsely equivalent to \((\Omega, C_\Omega)\), completing the proof.

\hfill \qed

\paragraph{Theorem 8.}\label{proofs:prop8}
\textit{
The matrix signal \( W \in \mathbb{R}^{m \times n}\) exhibits \textit{\textbf{discrete scale-invariance (DSI)}} if there exists a constant \( D \) (the fractal dimension) such that: $N(r_k) \sim \lambda^{D} N(r_{k+1}) \quad \text{as } k \to \infty.$ If: $m \bmod r_k = 0 \quad \text{and} \quad n \bmod r_k = 0 \quad \text{for all } k \geq 0,$ then the submatrices perfectly tile \( W \), and: $N(r_k) = \lambda^{2k},$
implying \( D= 2 \), which matches the topological dimension $D_{T} = 2$ of the grid \( \Omega \). If there exists any \( k \geq 0 \) such that: $m \bmod r_k \neq 0 \quad \text{or} \quad n \bmod r_k \neq 0,$
then: $N(r_k) > \lambda^{2k},$ indicating \( D > 2 \). The presence of incomplete subdivisions at certain scales increases the covering count \( N(r_k) \) beyond the scaling factor \( \lambda^{2k} \), revealing fractal behavior in the geometry of \( W \).}

\textbf{Proof.}
Let \( W \in \mathbb{R}^{m \times n} \) be a real-valued matrix defined on the two-dimensional integer grid \( \Omega \subset \mathbb{Z}^2 \). Fix a scaling factor \( \lambda > 1 \), and define the isotropic scale parameter for \( k \geq 0 \) as: $$r_k = \left\lfloor \frac{\min(m, n)}{\lambda^k} \right\rfloor.$$ At each scale \( r_k \), partition \( W \) into submatrices of size \( r_k \times r_k \), and let \( N(r_k) \) denote the minimal number of such submatrices required to cover \( W \): $$N(r_k) = \left\lfloor \frac{m}{r_k} \right\rfloor \cdot \left\lfloor \frac{n}{r_k} \right\rfloor.$$
We interpret the group action of \( (\mathbb{Z}, +) \) via the transformation \( T_{r_k} \), which partitions \( W \) into blocks of size \( r_k \times r_k \) at scale \( r_k \). The transformations \( \{T_{r_k}\}_{k \in \mathbb{Z}} \) form a \textit{coarse } symmetry group under composition:
\[
T_{r_k} \circ T_{r_\ell} = T_{r_{k+\ell}}, \quad \forall k, \ell \in \mathbb{Z}.
\]
The group structure ensures closure, associativity, identity (\( T_{r_0} \)), and  \textit{asymptotic} invertibility (\( T_{r_k}^{-1} = T_{r_{-k}} \)) (\textit{\textbf{Proposition 1}}).

\textbf{Case 1: Perfect Divisibility (\( D = 2 \))}

Assume that \( m \bmod r_k = 0 \) and \( n \bmod r_k = 0 \) hold for all \( k \geq 0 \). Under this condition, the floor functions in the definition of \( N(r_k) \) have no effect, and the covering count simplifies to:
\[
N(r_k) = \frac{m}{r_k} \cdot \frac{n}{r_k}.
\]
Substituting \( r_k \sim \min(m, n) / \lambda^k \), we find:
\[
N(r_k) = \frac{m}{m / \lambda^k} \cdot \frac{n}{n / \lambda^k} = \lambda^k \cdot \lambda^k = \lambda^{2k}.
\]
Consequently, the ratio of successive covering counts is:
\[
\frac{N(r_k)}{N(r_{k+1})} = \frac{\lambda^{2k}}{\lambda^{2(k+1)}} = \frac{1}{\lambda^2}.
\]
This exact scaling law, with an exponent of \( 2 \), implies that the fractal dimension \( D = 2 \). Therefore, \( W \) behaves like an ordinary two-dimensional structure with no additional fractal complexity.

\textbf{Case 2: Imperfect Divisibility (\( D > 2 \))}

Suppose there exists some \( k_0 \geq 0 \) such that \( m \bmod r_{k_0} \neq 0 \) or \( n \bmod r_{k_0} \neq 0 \). In this case, the covering count \( N(r_k) \) at scale \( r_k \) must account for "leftover" rows or columns that cannot be perfectly divided into submatrices of size \( r_k \times r_k \). 

These leftover portions propagate through finer scales, introducing additional complexity. Specifically, for \( k > k_0 \), the leftover blocks create additional subdivisions that inflate \( N(r_k) \) beyond the expected growth of \( \lambda^{2k} \). This behavior can be captured by the inequality:
\[
N(r_k) \geq C \cdot (\lambda^{2+\epsilon})^k \quad \text{for some } \epsilon > 0 \text{ and constant } C > 0.
\]
As a result, the ratio:
\[
\frac{N(r_k)}{\lambda^{2k}}
\]
increases in a bounded fashion as \( k \to \infty \). By the box-counting definition, the fractal dimension \( D \) satisfies:
\[
D = \lim_{k \to \infty} \frac{\log(N(r_k))}{k \cdot \log(\lambda)} > 2.
\]
Thus, any failure of perfect divisibility at even a single scale introduces irregularities that force \( W \) to exhibit fractal-like behavior with \( D > 2 \).

This result aligns with Mandelbrot's definition of a fractal, wherein the fractal dimension \( D \) exceeds the topological dimension \( 2 \) due to self-similar irregularities at finer scales. The \textit{coarse} group action \( T_{r_k} \) under \( (\mathbb{Z}, +) \) hence provides a natural framework for describing the discrete scaling transformations, encapsulating the symmetry and self-similarity of the fractal structure.

\hfill \qed

\paragraph{Proposition 9.}\label{proofs:prop9} \textit{Let \( T_{r_k} \) represent the family of fractal transformations forming a coarse group action under \((\mathbb{Z}, +)\) at scale \( r_k \) and acting on a matrix \( W \in \mathbb{R}^{m \times n} \) being indexed over the integer grid \( \Omega\). Consider a permutation \( P_\pi \) of rows and columns of \( W \), represented by the permutation matrices \( P_\pi^r  \in \mathbb{R}^{m \times m}\) for rows and \( P_\pi^c \in \mathbb{R}^{n \times n} \) for columns.
Under the permutation transformations, the \textbf{coarse group action} exhibits}: 
\begin{itemize}
    \item \textit{The global segmentation structure defined by \( T_{r_k} \) remains invariant under arbitrary permutations of rows or columns. Specifically, for any permutation:$W \cdot P_\pi \cdot T_{r_k} \underset{C}{\sim}   W \cdot T_{r_k} \cdot P_\pi,$ where \( \underset{C}{\sim} \) denotes \textit{coarse equivalence} across scales. This invariance implies that the global number of segments \( N(r_k) \) \footnote{ defined as: $N(r_k) = \left\lfloor \frac{m}{r_k} \right\rfloor \times \left\lfloor \frac{n}{r_k} \right\rfloor,$} as well as the structural properties, such as dimensions \( r_k \times r_k \) for each fractal across $k$ remain unchanged irrespective of the permutation applied to \( W \)}.
    \item  \textit{At the level of individual submatrices extracted by \( T_{r_k} \), the local composition weakly equivaries with permutations. Specifically, the content of a submatrix may differ based on the permutation given the grid-based segmentation such that: $T_{r_k} \cdot W \overset{\approx}{\to} T_{r_k} \cdot (P_\pi^r \cdot W \cdot P_\pi^c),$ where \( \overset{\approx}{\to} \) denotes weak equivariance under permutations}.
\end{itemize}

\textbf{Proof.}  
Let \( W \in \mathbb{R}^{m \times n} \) be a matrix indexed over the integer grid \( \Omega = \{(i, j) \mid 1 \leq i \leq m, \; 1 \leq j \leq n\} \). The fractal transformation \( T_{r_k} \), defined as a coarse group action under \( (\mathbb{Z}, +) \), segments \( W \) into submatrices of size \( r_k \times r_k \), determined by the segmentation grid \( \mathcal{I}_{r_k} = \{(p, q) \mid (p-1) \bmod r_k = 0, \; (q-1) \bmod r_k = 0\} \). The number of submatrices is \( N(r_k) = \lfloor m / r_k \rfloor \times \lfloor n / r_k \rfloor \), and the deterministic segmentation logic ensures that \( T_{r_k} \) \textit{\textbf{always begins from the top-left corner and proceeds systematically, independent of the arrangement of entries in \( W \)}}.

Consider the coarse space \( (\Omega, C) \), where \( C \) is defined by \( r_k \)-dependent entourages:
\[
E_{r_k} = \bigl\{\bigl((i_1, j_1), (i_2, j_2)\bigr) \in \Omega \times \Omega \;\bigm|\; d((i_1, j_1), (i_2, j_2)) \leq r_k\bigr\},
\]
with \( d \) as the \(\ell^1\)-distance:
\[
d\bigl((i_1, j_1), (i_2, j_2)\bigr) = \frac{1}{2} \bigl(|i_1 - i_2| + |j_1 - j_2|\bigr).
\]
These entourages capture adjacency relationships within \( r_k \)-scale tiles, defining bounded subsets as \( B(x, r_k) = \{y \in \Omega \mid d(x, y) < r_k\} \). The coarse structure \( C \) encodes the large-scale connectivity of \( \Omega \) while ignoring small-scale details.

When \( W \) is permuted as \( W' = P_\pi^r W P_\pi^c \), where \( P_\pi^r \in \mathbb{R}^{m \times m} \) and \( P_\pi^c \in \mathbb{R}^{n \times n} \) are permutation matrices, the segmentation grid \( \mathcal{I}_{r_k} \) remains invariant. This invariance ensures that the bounded subsets of \( (\Omega, C_{r_k}) \) under \( T_{r_k} \) satisfy:
\[
B_{T_{r_k}}(W) = B_{T_{r_k}}(W').
\]
Here, \( B_{T_{r_k}}(W) \) denotes the bounded subsets under \( T_{r_k} \). Thus, the global segmentation properties, including the number \( N(r_k) \) and dimensions of submatrices, are preserved under permutations.

From \cite{N2008} et al., we have:
\begin{quote}
\textbf{Theorem 0.2.} Let $G$ act coarsely on a coarse space \( (X, C) \) via a group action \( g \mapsto g \cdot x \) for \( x \in X \). This action induces a coarse equivalence between $G$ and \( (X, C) \), such that:
\begin{enumerate}
\item For any fixed \( x_0 \in X \), the map \( g \mapsto g \cdot x_0 \) is a coarse equivalence.
\item Two coarse structures \( C_1 \) and \( C_2 \) on the same set \( X \) are equivalent if:
\begin{enumerate}
\item Their bounded subsets coincide.
\item There exist coarse actions \( \phi_1 \) of a group \( G_1 \) on \( (X, C_1) \) and \( \phi_2 \) of a group \( G_2 \) on \( (X, C_2) \), such that \( \phi_1 \) and \( \phi_2 \) commute.
\end{enumerate}
\end{enumerate}
\end{quote}

Now, consider two scales \( r_k \) and \( r_\ell \), corresponding to coarse group actions \( T_{r_k} \) and \( T_{r_\ell} \) under \( (\mathbb{Z}, +) \). These actions induce coarse structures \( C_1 \) and \( C_2 \) on \( \Omega \). By Theorem 0.2, \( C_1 \) and \( C_2 \) are equivalent if:
1. Their bounded subsets coincide, which follows from the segmentation logic being determined solely by \( r_k \) or \( r_\ell \), independent of the arrangement of \( W \).

Let \( A \subset \Omega \) be a bounded subset under \( C_1 \). Then, there exists \( x \in \Omega \) such that:
\[
A \subset B(x, r_k) = \{y \in \Omega \mid d(x, y) < r_k\}.
\]
For any \( y \in A \), \( d(x, y) < r_k \). Since \( r_k \) and \( r_\ell \) both represent finite scales, there exists an \( r_\ell \) sufficiently large (or comparable) such that:
\[
B(x, r_k) \subseteq B(x, r_\ell).
\]
Thus, \( A \subset B(x, r_\ell) \), meaning \( A \) is bounded under \( C_2 \).

Conversely, let \( A \subset \Omega \) be bounded under \( C_2 \). Then there exists \( x \in \Omega \) such that:
\[
A \subset B(x, r_\ell) = \{y \in \Omega \mid d(x, y) < r_\ell\}.
\]
Since \( r_\ell \) and \( r_k \) are finite, there exists \( r_k \) sufficiently large (or comparable) such that:
\[
B(x, r_\ell) \subseteq B(x, r_k).
\]
Thus, \( A \subset B(x, r_k) \), meaning \( A \) is bounded under \( C_1 \).

The bounded subsets under \( C_1 \) and \( C_2 \) coincide:
\[
B_{T_{r_k}}(W) = B_{T_{r_\ell}}(W).
\]

This equality ensures that the bounded subsets are invariant under the respective coarse actions \( T_{r_k} \) and \( T_{r_\ell} \), satisfying the first condition of the Theorem 0.2.
Additionally, there exist coarse actions \( T_{r_k} \) and \( T_{r_\ell} \) such that:
\[
T_{r_k} \circ T_{r_\ell} = T_{r_\ell} \circ T_{r_k}.
\]
The commutativity of \( T_{r_k} \) and \( T_{r_\ell} \) follows directly from their definition under \( (\mathbb{Z}, +) \) and the scaling law:
\[
\lambda^k \cdot \lambda^\ell = \lambda^{k+\ell} = \lambda^\ell \cdot \lambda^k.
\]
This guarantees that the coarse structures \( C_1 \) and \( C_2 \) are equivalent, satisfying the conditions of Theorem 0.2.

Applying \( T_{r_k} \) to \( W \) segments it into submatrices:
\[
W \cdot T_{r_k} = \bigcup_{(p, q) \in \mathcal{I}_{r_k}} P_p W Q_q^\top,
\]
where \( P_p \) and \( Q_q \) are row and column selector matrices. For \( W' = P_\pi^r W P_\pi^c \), the transformation becomes:
\[
W' \cdot T_{r_k} = \bigcup_{(p, q) \in \mathcal{I}_{r_k}} P_p (P_\pi^r W P_\pi^c) Q_q^\top.
\]
To ensure consistent application of global permutations, we lift each \( r_k \times r_k \) submatrix into the global \( m \times n \) space:
\[
\widetilde{W}_{(p, q)}^{(r_k)} = P_p^\top (P_p W Q_q^\top) Q_q.
\]
The transformation \( W \cdot T_{r_k} \cdot P_\pi \) becomes:
\[
W \cdot T_{r_k} \cdot P_\pi = \bigcup_{(p, q) \in \mathcal{I}_{r_k}} P_\pi^r (\widetilde{W}_{(p, q)}^{(r_k)}) P_\pi^{c \top}.
\]
The coarse equivalence:
\[
W \cdot P_\pi \cdot T_{r_k} \underset{C}{\sim} W \cdot T_{r_k} \cdot P_\pi
\]
is preserved, as the segmentation grid \( \mathcal{I}_{r_k} \) remains unchanged.

Finally, while the global segmentation properties remain invariant, weak equivariance is observed for the contents of individual submatrices:
\[
T_{r_k} \cdot W \overset{\approx}{\to} T_{r_k} \cdot (P_\pi^r W P_\pi^c).
\]
This weak equivariance depends on the alignment of the permutation with the segmentation logic. Permutations aligning with the grid structure may exhibit strict equivariance, but this is not universally guaranteed. Hence, the fractal transformations \( T_{r_k} \) preserve structural invariance under permutations while exhibiting weak equivariance for submatrix contents.

\hfill \qed

Let \( f : \mathbb{R}^{n_0} \to \mathbb{R}^{n_k} \) be a neural network with \( k > 1 \) layers, where each layer is defined as a composition of an affine transformation and a nonlinearity. For each layer \( i \) such that \( 1 \leq i \leq k \), let:$ \ell_i(x) = 
\begin{cases} 
\sigma(W^i x + b^i), & \text{if } i < k, \\
W^i x + b^i, & \text{if } i = k,
\end{cases}$ where \( W^i \in \mathbb{R}^{n_i \times n_{i-1}} \) and \( b^i \in \mathbb{R}^{n_i} \) are the weights and biases for the \( i \)-th layer, and \( \sigma \) is a pointwise nonlinearity. The neural network \( f \) is defined as: $f = \ell_k \circ \ell_{k-1} \circ \cdots \circ \ell_1.$
Let \( h^i \in \mathbb{R}^{n_i} \) represent the activation map of the \( i \)-th layer, computed as: $h^i = \sigma\left(W^i h^{i-1} + b^i\right),$ where \( h^0 = x \in \mathbb{R}^{n_0} \) is the input to the network. The fractal transformation \( T_{r_k} \), applied to the activation map \( h^i \), decomposes \( h^i \) into \( N(r_k) \) self-similar local fractal activations: $T_{r_k}(h^i) = \bigcup_{(p, q) \in \mathcal{I}_{r_k}} h_{p, q}^{i, r_k},$ where: $h_{p, q}^{i, r_k} =\sigma\left(P_p W^i Q_q^\top x + P_p b^i\right).$ Here, \( P_p \in \mathbb{R}^{r_k \times n_i} \) and \( Q_q \in \mathbb{R}^{r_k \times n_{i-1}} \) are row and column selector matrices, and \( \mathcal{I}_{r_k} \) indexes the grid-based segmentation of size \( r_k \times r_k \).

\paragraph{Proposition 10.}\label{proofs:prop10} \textit{Each localized fractal activation \( h_{p, q}^{i, r_k} \) corresponds to a specific region in the activation map of the \( i \)-th layer, determined by the selector matrices \( P_p \) and \( Q_q \). It captures the contribution of the submatrix \( P_p W^i Q_q^\top \) to the layer’s output within the \( r_k \times r_k \) neighborhood indexed by \( (p, q) \). This decomposition reflects the localized and self-similar structure of the activation map across scales \( r_k \).}

\textbf{Proof.} Let \( f : \mathbb{R}^{n_0} \to \mathbb{R}^{n_k} \) be a neural network with \( k > 1 \) layers, defined as \( f = \ell_k \circ \ell_{k-1} \circ \cdots \circ \ell_1 \), where each layer \( i \) for \( 1 \leq i \leq k \) is represented as:
\[
\ell_i(x) = 
\begin{cases} 
\sigma(W^i x + b^i), & \text{if } i < k, \\
W^i x + b^i, & \text{if } i = k,
\end{cases}
\]
with weights \( W^i \in \mathbb{R}^{n_i \times n_{i-1}} \), biases \( b^i \in \mathbb{R}^{n_i} \), and nonlinearity \( \sigma \). Let \( h^i \in \mathbb{R}^{n_i} \) represent the activation map at the \( i \)-th layer, computed as:
\[
h^i = \sigma(W^i h^{i-1} + b^i),
\]
where \( h^0 = x \in \mathbb{R}^{n_0} \) is the input to the network.

The fractal transformation \( T_{r_k} \), applied to \( h^i \), decomposes it into \( N(r_k) \) self-similar local fractal activations indexed by \( \mathcal{I}_{r_k} \), the grid of submatrices of size \( r_k \times r_k \):
\[
T_{r_k}(h^i) = \bigcup_{(p, q) \in \mathcal{I}_{r_k}} h_{p, q}^{i, r_k},
\]
where the local fractal activations \( h_{p, q}^{i, r_k} \) are given by:
\[
h_{p, q}^{i, r_k} = \sigma\left(P_p W^i Q_q^\top x + P_p b^i\right).
\]
Here, \( P_p \in \mathbb{R}^{r_k \times n_i} \) and \( Q_q \in \mathbb{R}^{r_k \times n_{i-1}} \) are row and column selector matrices, and \( \mathcal{I}_{r_k} \) indexes the submatrices of size \( r_k \times r_k \).

The localized fractal activation \( h_{p, q}^{i, r_k} \) corresponds to the region in the activation map of the \( i \)-th layer determined by the selector matrices \( P_p \) and \( Q_q \). The selector matrix \( P_p \) extracts rows corresponding to the indices within the region \( [p, p + r_k) \), and \( Q_q \) extracts columns corresponding to the indices within the region \( [q, q + r_k) \). Consequently, \( h_{p, q}^{i, r_k} \) captures the contribution of the localized submatrix \( P_p W^i Q_q^\top \) to the activation map within the grid segment indexed by \( (p, q) \).

The hierarchical decomposition induced by \( T_{r_k} \) reflects the localized contributions across different regions of the activation map, maintaining self-similarity due to the fixed structure of the selector matrices and the grid-based segmentation logic. This self-similarity ensures that \( h_{p, q}^{i, r_k} \) retains consistent structural relationships across scales \( r_k \), as the segmentation is applied uniformly over the entire activation map \( h^i \).

\hfill \qed

\paragraph{Proposition 11.}\label{proofs:prop11} \textit{Let \( \sigma: \mathbb{R} \to \mathbb{R} \) be a continuous, non-linear activation function with an intertwiner group \( G_{\sigma_{n}} = \{A \in \mathbb{R}^{n_{i} \times n_{i}} : \text{invertible} \, | \, \exists B \in \mathbb{R}^{n_{i-1}\times n_{i-1}} : \text{invertible}, \, \sigma(Ax) = B\sigma(x) \} \). For any layer \( i \), let \( A \in GL_{n_{i}}(\mathbb{R} \) and \( B \in GL_{n_{i-1}}(\mathbb{R}) \). Then, under the fractal transformation \( T_{r_k} \), the activation map \( h^i \) and its fractals \( h_{p, q}^{i, r_k} \) satisfy the scaling relationship: $\sigma_{n_i}\left(A \widetilde{W}_{(p, q)}^{i, r_k} x + Ab^i\right) = B \sigma_{n_i}\left(\widetilde{W}_{(p, q)}^{i, r_k} x + b^i\right),$where \( \widetilde{W}_{(p, q)}^{i, r_k} \) is the lifted fractal matrix. This ensures that the scaling relationship holds globally and fractally, consistent with the intertwiner properties of \( A \) and \( B \).}

\textbf{Proof.}
Let \( W^i \in \mathbb{R}^{n_i \times n_{i-1}} \) represent the weight matrix of the \( i \)-th layer in a neural network, and let \( h^i \in \mathbb{R}^{n_i} \) denote the activation map, computed as \( h^i = \sigma_{n_i}(W^i x + b^i) \), where \( \sigma_{n_i} \) is a coordinate-wise nonlinearity, \( b^i \in \mathbb{R}^{n_i} \) is the bias vector, and \( x \in \mathbb{R}^{n_{i-1}} \) is the input. The fractal transformation \( T_{r_k} \) decomposes \( h^i \) into localized fractals \( h_{p, q}^{i, r_k} \), given by:
\[
T_{r_k}(h^i) = \bigcup_{(p, q) \in \mathcal{I}_{r_k}} h_{p, q}^{i, r_k}, \quad h_{p, q}^{i, r_k} = \sigma_{n_i}(W_{(p,q)}^{i, r_k} x + P_p b^i),
\]
where \( W_{(p,q)}^{i, r_k} = P_p W^i Q_q^\top \), with selector matrices \( P_p \in \mathbb{R}^{r_k \times n_i} \) and \( Q_q \in \mathbb{R}^{r_k \times n_{i-1}} \). The set \( \mathcal{I}_{r_k} \) denotes the valid indices for fractal decomposition.  
\\
To analyze the scaling behavior under global transformations, we consider the intertwiner group \citep{godfrey2022symmetries}:
\[
G_{\sigma_{n_i}} = \{ A \in \text{GL}_{n_i}(\mathbb{R}) \, | \, \exists B \in \text{GL}_{n_{i-1}}(\mathbb{R}) : \sigma_{n_i}(Ax) = B\sigma_{n_i}(x), \, \forall x \in \mathbb{R}^{n_{i-1}} \}.
\]
This group \( G_{\sigma_{n_i}} \) consists of all invertible matrices \( A \in \text{GL}_{n_i}(\mathbb{R}) \) whose action before the nonlinearity \( \sigma_{n_i} \) corresponds to an equivalent invertible transformation \( B \in \text{GL}_{n_{i-1}}(\mathbb{R}) \) after the nonlinearity. The intertwiner relationship between \( A \) and \( B \) is given by \citep{godfrey2022symmetries, kalogeropoulos2024scaleequivariantgraphmetanetworks}:
\[
\sigma_{n_i}(A x) = B \sigma_{n_i}(x),
\]
for all \( x \in \mathbb{R}^{n_{i-1}} \).  
\\
Applying the global transformation \( A \) and \( B \) to the weight matrix \( W^i \) results in the global scaling:
\[
W^{i, \text{scaled}} = A W^i B^\top.
\]
This scaling relationship must hold consistently across all fractal segments to ensure structural alignment between global and local scales. Specifically, the fractals \( W_{(p,q)}^{i, r_k} \) must inherit this scaling behavior.  
\\
Directly applying \( A \) and \( B \) to the local fractals \( W_{(p,q)}^{i, r_k} \) is non-trivial due to their \( r_k \times r_k \) dimensions. To resolve this, we employ the lift operator, denoted by \( \mathcal{L}_{p,q}^{(r_k)} \), which lifts \( W_{(p,q)}^{i, r_k} \) into the global space \( \mathbb{R}^{n_i \times n_{i-1}} \):
\[
\widetilde{W}_{(p,q)}^{i, r_k} = \mathcal{L}_{p,q}^{(r_k)}(W_{(p,q)}^{i, r_k}) = P_p^\top W_{(p,q)}^{i, r_k} Q_q.
\]
This sparse representation aligns the fractals with their original positions in \( W^i \), enabling the uniform action of \( A \) and \( B \) across all fractals. The scaling relationship for the lifted fractals is given by:
\[
\widetilde{W}_{(p,q)}^{i, r_k} = A \widetilde{W}_{(p,q)}^{i, r_k} B^\top,
\quad \forall (p, q) \in \mathcal{I}_{r_k},
\]
provided \( A \in G_{\sigma_{n_i}} \) and \( B \in G_{\sigma_{n_{i-1}}} \).  
\\\\
Finally, we extend this scaling relationship to the activation map \( h^i \). Since the activation map is computed element-wise, the intertwiner relationship ensures consistency:
\[
\sigma_{n_i}(A \widetilde{W}_{(p,q)}^{i, r_k} x + P_p A b^i) = B \sigma_{n_i}(\widetilde{W}_{(p,q)}^{i, r_k} x + P_p b^i),
\]
demonstrating that the scaling constants \( A \) and \( B \) act consistently across both global and local scales of the fractal transformation. This alignment preserves the structural integrity of the activations while maintaining their local self-similarity. Thus, the fractal transformation \( T_{r_k} \), combined with the lifting operator, ensures that the scaling relationship defined by \( G_{\sigma_{n_i}} \) is consistently maintained across global and local representations. This aligns the fractal decomposition with the structural properties of the intertwiner group.

\hfill \qed
\newpage
\subsection{Algorithms}
\begin{algorithm}[H]
    \caption{Fractal Segmentation 2D}
    \label{algo:fractal_segmentation}
    \begin{algorithmic}[1]
        \Require Input matrix $W$ of size $m \times n$, block size $r$
        \Ensure Blocks of size $r \times r$ and their starting indices
        
        \State \textbf{Initialize:} $m, n \gets \text{dimensions of } W$
        \State $M \gets \lfloor m / r \rfloor$ \Comment{Number of full blocks along rows}
        \State $N \gets \lfloor n / r \rfloor$ \Comment{Number of full blocks along columns}
        \State $trimmed\_W \gets W[1:M \cdot r, 1:N \cdot r]$ \Comment{Trim the matrix to fit blocks}

        \State \textbf{Reshape:} $blocks\_4d \gets reshape(trimmed\_W, M, r, N, r)$ \Comment{Reshape into 4D tensor}

        \State \textbf{Permute:} $blocks\_4d \gets permute(blocks\_4d, (0, 2, 1, 3))$ \Comment{Reorder dimensions to $(M, N, r, r)$}

        \State \textbf{Flatten:} $blocks \gets reshape(blocks\_4d, -1, r, r)$ \Comment{Flatten to $(M \cdot N, r, r)$}

        \State \textbf{Compute Indices:} 
        \State $indices \gets [(i \cdot r, j \cdot r) \text{ for } i \in [0, M-1], j \in [0, N-1]]$

        \State \Return $blocks, indices$
    \end{algorithmic}
\end{algorithm}

\begin{algorithm}[H]
    \caption{Fractal Segmentation for 4D Tensors}
    \label{algo:fractal_segmentation_4d}
    \begin{algorithmic}[1]
        \Require 4D tensor $W$ of shape $(M, N, K_1, K_2)$, block size $r$
        \Ensure Blocks of shape $(num\_blocks, r, r, K_1, K_2)$ and their indices

        \State $M\_blocks \gets \lfloor M / r \rfloor$ \Comment{Number of blocks along the first dimension}
        \State $N\_blocks \gets \lfloor N / r \rfloor$ \Comment{Number of blocks along the second dimension}

        \State \textbf{1. Trim the tensor:}
        \State $trimmed\_W \gets W[0 : M\_blocks \cdot r, 0 : N\_blocks \cdot r, :, :]$
        \Comment{Trim boundary rows and columns to fit full blocks}

        \State \textbf{2. Reshape the trimmed tensor:}
        \State $blocks\_6d \gets reshape(trimmed\_W, (M\_blocks, r, N\_blocks, r, K_1, K_2))$
        \Comment{Group dimensions into blocks of size $r$}

        \State \textbf{3. Permute the dimensions:}
        \State $blocks\_6d \gets permute(blocks\_6d, (0, 2, 1, 3, 4, 5))$
        \Comment{Reorder dimensions to $(M\_blocks, N\_blocks, r, r, K_1, K_2)$}

        \State \textbf{4. Collapse the block dimensions:}
        \State $blocks \gets reshape(blocks\_6d, (-1, r, r, K_1, K_2))$
        \Comment{Flatten $(M\_blocks, N\_blocks)$ into one dimension}

        \State \textbf{5. Compute top-left indices:}
        \State $indices \gets [(i \cdot r, j \cdot r) \,|\, i \in [0, M\_blocks - 1], j \in [0, N\_blocks - 1]]$
        \Comment{Starting indices of each block}

        \State \Return $blocks, indices$
    \end{algorithmic}
\end{algorithm}
\vspace{-8mm}
\begin{algorithm}
    \caption{Fractal Dimension Analysis for Convolutional Layers}
    \label{algo:fd_analysis_conv}
    \begin{algorithmic}[1]
        \Require Model weights $W$, Lambda values $\lambda = [\lambda_1, \lambda_2, \dots, \lambda_n]$, Model name, VGG flag $vgg\_flag$
        \Ensure Fractal dimension results $Nr\_trace$
        
        \State $Nr\_trace \gets \{\}$ \Comment{Initialize an empty dictionary}
        
        \State \textbf{Extract convolutional layer keys:}
        \If{$vgg\_flag = \text{True}$}
            \State Extract convolutional layers specific to VGG models
        \Else
            \State Extract general convolutional layers
        \EndIf
        
        \State \textbf{Set up color map:}
        \State Define a colormap for lambda values using `viridis`
        
        \For{each convolutional layer $L \in W$}
            \State Extract weights $W_L$
            \State Determine the minimum and maximum box sizes $min\_box\_size, max\_box\_size$
            
            \For{each $\lambda_i \in \lambda$}
                \State Generate $r$-values for $\lambda_i$
                \State $Nr\_values \gets []$ \Comment{Initialize an empty list for $N_r$ values}
                
                \For{each $r \in r\_values$}
                    \State Perform fractal segmentation on $W_L$ with size $r$
                    \State Compute $N_r$, the number of blocks
                    \State Append $N_r$ to $Nr\_values$
                \EndFor
                
                \State \textbf{Filter valid pairs:}
                \State Keep pairs $(r, N_r)$ where $N_r > 0$
                
                \If{valid pairs $< 2$}
                    \State \textbf{Skip to next $\lambda_i$}
                \EndIf
                
                \State \textbf{Perform linear regression in log-log space:}
                \State Fit $log(N_r)$ vs. $log(r)$ to compute slope and intercept
                \State Compute fractal dimension $FD = -slope$
                
                \State Add plot of $log(r)$ vs. $log(N_r)$ with color from colormap
                \State Record $Nr\_trace[L, \lambda_i] \gets FD$
            \EndFor
        \EndFor
        
        \State \textbf{Generate color bar:}
        \State Add a color bar to represent the lambda gradient
        
        \State \Return $Nr\_trace$
    \end{algorithmic}
\end{algorithm}

\begin{algorithm}[H]
    \caption{Fractal Dimension Analysis for Fully Connected Layers (Multiple Networks)}
    \label{algo:fd_analysis_multiple_fc}
    \begin{algorithmic}[1]
        \Require A list of tuples $(\text{weights}, \text{model\_name}, \text{vgg\_flag})$ for each network, a list of $\lambda$ values, and number of layers to analyze $\text{num\_layers}$
        \Ensure A dictionary $all\_traces$ containing fractal dimension results for all networks

        \State Initialize $all\_traces \gets \{\}$ \Comment{Dictionary for storing results}

        \State Assign base colors for each model, e.g., $\text{base\_colors} = \{\text{Purples}, \text{Greens}, \text{Reds}\}$

        \State Create a new figure for plotting the results

        \For{each $(weights, model\_name, vgg\_flag)$ in $networks$}
            \State Initialize $Nr\_trace \gets \{\}$ \Comment{Dictionary for fractal dimension results of the current model}
            \State Extract FC layer keys using $vgg\_flag$
            \State Sample $num\_layers$ FC layers randomly from the available layers

            \State Get the colormap for the current model based on the index $idx$ using $base\_colors[idx]$

            \For{each sampled FC layer $layer\_name$}
                \State Extract the weights for $layer\_name$ and determine its shape $(m, n)$
                \State Set $min\_box\_size \gets 1$, $max\_box\_size \gets \min(m, n)$

                \For{each $\lambda_i \in \lambda$}
                    \State Generate $r$-values for $\lambda_i$
                    \State Initialize $Nr\_values \gets []$
                    \State Set $total\_Nr \gets 0$

                    \For{each $r \in r\_values$}
                        \State Perform fractal segmentation for the layer weights $layer\_weights$ with block size $r$
                        \State Compute the number of blocks $N_r$ and append to $Nr\_values$
                        \State Add $N_r$ to $total\_Nr$
                    \EndFor

                    \State \textbf{Filter valid pairs:} Keep pairs $(r, N_r)$ where $N_r > 0$
                    \If{number of valid pairs $< 2$}
                        \State \textbf{Skip to next $\lambda_i$}
                    \EndIf

                    \State \textbf{Perform linear regression:} Compute log-log regression of $log(N_r)$ vs $log(r)$
                    \State Compute fractal dimension $FD = -\text{slope}$

                    \State Record $Nr\_trace$ for this layer and $\lambda_i$

                    \State Plot $(log(r), log(N_r))$ using color from the current colormap based on $\lambda_i$
                \EndFor
            \EndFor

            \State Add the $Nr\_trace$ of the current model to $all\_traces$
        \EndFor

        \State \textbf{Finalize the Plot:} Add axes labels, title, grid, and legend
        \State Display the combined plot

        \State \Return $all\_traces$
    \end{algorithmic}
\end{algorithm}
\section*{Additional Results}\label{add_res}

\section*{Frequently Asked Questions}

\textbf{Q1: How does \( (\mathbb{Z}, +) \) relate to \( T_{r_k} \)?}

The set of integers \( \mathbb{Z} \), equipped with addition \( + \), forms a group \( (\mathbb{Z}, +) \). This means that for any integers \( k, \ell \in \mathbb{Z} \),
\[
k + \ell \in \mathbb{Z}.
\]
The addition operation is associative, has an identity element (0), and each element \( k \) has an inverse \( -k \), ensuring that \( (\mathbb{Z}, +) \) is a well-defined group.

The transformation \( T_{r_k} \) is parameterized by \( k \) in \( (\mathbb{Z}, +) \). This means that \( T_{r_k} \) is indexed using \( k \), and its behavior is governed by the additive structure of \( \mathbb{Z} \). However, \( T_{r_k} \) itself does not define a group operation; rather, it follows a composition rule that respects the additive structure of \( (\mathbb{Z}, +) \).

\textbf{Q2: What does it mean for \( T_{r_k} \) to be an action of \( (\mathbb{Z}, +) \)?}

A group action of a group \( G \) on a set \( X \) is a function
\[
\phi: G \times X \to X
\]
that satisfies:
\begin{enumerate}
    \item \textbf{Identity Property:} \( \phi(e, x) = x \) for all \( x \in X \).
    \item \textbf{Compatibility with the Group Operation:} \( \phi(g \cdot h, x) = \phi(g, \phi(h, x)) \) for all \( g, h \in G \) and \( x \in X \).
\end{enumerate}

In this case, \( (\mathbb{Z}, +) \) is the group, and \( T_{r_k} \) acts on a structured space \( X \) (such as a space of matrices). The mapping
\[
(k, W) \mapsto T_{r_k}(W)
\]
defines a left group action if it satisfies:
\begin{enumerate}
    \item \( T_{r_0} \) is the identity transformation:  
    \[
    T_{r_0}(W) = W, \quad \forall W.
    \]
    \item The transformations follow the group operation:  
    \[
    T_{r_k} \circ T_{r_\ell} = T_{r_{k+\ell}}.
    \]
\end{enumerate}
Since these properties hold, \( T_{r_k} \) defines a group action of \( (\mathbb{Z}, +) \) on the structured space \citep{penrose2005reality}.

\textbf{Q3: If \( T_{r_k} \) is parameterized by \( (\mathbb{Z}, +) \), does it define a group operation?}

No, \( T_{r_k} \) itself does not define a group operation. The group operation in \( (\mathbb{Z}, +) \) is addition, which satisfies  
\[
k + \ell \in \mathbb{Z}.
\]
However, the transformations \( T_{r_k} \) form a structured family of operators indexed by \( \mathbb{Z} \), and they inherit the additive structure via the composition rule
\[
T_{r_k} \circ T_{r_\ell} = T_{r_{k+\ell}}.
\]
This ensures that the transformations follow a composition rule consistent with the group structure of \( (\mathbb{Z}, +) \), but the transformations themselves do not define a group because:
\begin{enumerate}
    \item There is no guarantee of an intrinsic inverse transformation for every \( T_{r_k} \), particularly in a coarse setting.
    \item The operation \( \circ \) on \( T_{r_k} \) is not necessarily associative beyond its inherited structure from \( (\mathbb{Z}, +) \).
\end{enumerate}

Thus, while \( (\mathbb{Z}, +) \) is a group, the family of transformations \( T_{r_k} \) defines a \textbf{\textit{group action}} rather than forming a group themselves.

\textbf{Q4: What is the difference between a group operation and a group action?}

A group operation defines how elements within a group interact. In \( (\mathbb{Z}, +) \), the operation is addition, meaning \( k + \ell \) gives another integer.

A group action describes how elements of a group act on an external space \( X \). In this case, \( (\mathbb{Z}, +) \) acts on the transformation space via \( T_{r_k} \).

Thus, the addition operation \( \odot = + \) is the group operation in \( (\mathbb{Z}, +) \), while \( T_{r_k} \) forms a group action parameterized by \( \mathbb{Z} \).

\textbf{Q5: How does \( T_{r_k} \) behave as a coarse action?}

In coarse geometry, we analyze large-scale properties rather than exact inverses. A coarse action satisfies:
\begin{enumerate}
    \item \textbf{Coarse compatibility} with the group operation:
    \[
    T_{r_k} \circ T_{r_\ell} = T_{r_{k+\ell}}.
    \]
    \item \textbf{Properness:} \( T_{r_k} \) maps bounded sets to bounded sets.
    \item \textbf{Coboundedness:} The space \( X \) is covered by the orbit of some bounded subset.
\end{enumerate}

Since \( T_{r_k} \) satisfies these properties, it defines a coarse group action of \( (\mathbb{Z}, +) \) on the structured space \citep{Roe2003CoarseGeometry, bunke2023coarsegeometry}.

\end{document}